# An AI-Based OSSE Framework to Evaluate Learned Physical Consistency: A Multi-Model Study Using FourCastNet, Aurora, Prithvi WxC, and GraphCast

**Yash Diggikar[1], Milton Halem[1], and Aishwarya Raman[2]**

[1]University of Maryland, Baltimore County, Baltimore, MD, USA

[2]Earth Sciences Division, NASA Ames Research Center, Moffett Field, CA, USA

Corresponding author: Yash Diggikar (yashd1@umbc.edu)

**Key Points:**

- First public AI OSSE temperature and wind inference impact under ERA5 reanalysis insertion on non-inserted variables: strongest for Prithvi WxC (52.2%), Aurora (46.2%), ForeCastNet (42.6%) and weakest for GraphCast (20.5%) of four AI models tested.
- Aurora and Prithvi WxC retain a positive tropical wind response to temperature insertion, confirming Charney's 1969 all-latitude conjecture.
- Forcing GraphCast with temperature truth at all 37 levels destabilizes the model; excluding 11 stratospheric levels at or below 100 hPa resolves the failure.

**Abstract**

Observing System Simulation Experiments (OSSEs) are a central methodology for evaluating future atmospheric observing systems. We present the first AI-based OSSE framework to test whether different AI weather models can reproduce comparable physical impact from new observations that traditional physics based models generate. We evaluate four distinct architectural AI models on the Charney conjecture that incomplete historical observations can infer the present state of the atmospheric. We considner– FourCastNet V2, GraphCast, Aurora and Prithvi WxC and integrate each autoregressively over a 60-day forecast horizon. After a 30-day spin-up, we insert ERA5 temperature fields to infer winds and related atmospheric fields. Similar OSSE experiments are conducted for wind insertion. We measure the performance of each model's response to observational information on a 30-day data assimilation cycle forecast with and without data insertion.

Distinct architectural AI weather models respond differently to the same OSSE observations. Foundation models (i.e. Prithvi WxC and Aurora), produce the strongest impact. For example, while GraphCast is the weakest, all three of the four models capture the temperature-to-geopotential height response. The two foundational models retain a positive, non-trivial, tropical zonal wind at 500 hPa response to temperature insertion. Various data insertion experiments can vary significantly even within a given architecture. We identify an architecture-specific numerical instability in GraphCast's full 37-level configuration that produces a runaway feedback concentrated at the uppermost levels with ERA5 temperature insertion correctable by excluding levels at or below 100 hPa.

**Plain Language Summary**

Weather forecasts increasingly rely on fast artificial-intelligence (AI) models instead of traditional physics-based computer models. Scientists want to know whether these AI models can also help decide where to place new weather satellites and instruments, a task that normally requires expensive physics-based simulations. We tested four differently built AI weather models by feeding each one perfect temperature, wind, or pressure information partway through a 60-day forecast, then checking whether the model used that single corrected piece of information to also fix its other, uncorrected predictions, the way real weather physics would. The models behaved very differently: two of them, built as general-purpose “foundation” models, spread the corrected information sensibly to other variables even in the tropics – a region where, for decades, scientists assumed this kind of correction mostly failed. This confirms a 1969 scientific prediction that had since fallen out of favor. A fourth, graph-based model struggled most, and in one setting became unstable until we adjusted which part of the atmosphere received the correction. These results suggest some AI weather models are already promising tools for cheaply testing future satellite and sensor designs, while others need more development before they can be trusted for this purpose.

## 1. Introduction

Recent advances in machine learning have produced global weather-forecast models such as FourCastNet V2 (Pathak et al., 2022; Bonev et al., 2023), Aurora (Bodnar et al., 2025), GraphCast (Lam et al., 2023), Prithvi WxC (Schmude et al., 2024) whose medium-range skill rivals or exceeds that of operational NWP while running orders-of-magnitude faster (Bi et al., 2023; Lam et al., 2023). Their speed makes them attractive for costly tasks such as Observing-System Simulation Experiments (OSSEs) (Hoffman & Atlas, 2016), yet their role inside the OSSE framework remains largely unexplored (Tian et al., 2024) because they omit the explicit data assimilation machinery that links observations to model states in conventional NWP, although dedicated physical-consistency evaluation frameworks for ML weather models have recently begun to emerge (Kasteleyn et al., 2026; Pasquini et al., 2026). Classical OSSEs use a high-resolution “nature run” to generate synthetic observations with realistic error characteristics into a data-assimilation system to quantify observing-system value. This approach has guided the design of satellite sounders, wind profilers, scatterometers, soil-moisture sensors, precipitation radars, aerosol lidars, and more (Atlas et al., 1985; Rohaly & Krishnamurti, 1993; Atlas, 1997; Zhang & Pu, 2010). In traditional NWP, physical equations, balance relationships (Holton & Hakim, 2013), and data assimilation (Kalnay, 2003; Daley, 1991) govern information transfer; in ML models, it occurs only through relationships learned from data.

### a. ML-based OSSE framework

We adapt classical OSSE logic to AI weather models through a simplified direct-insertion ML-OSSE framework. Rather than simulating observations and cycling them through a complete data-assimilation system, such as 4D-Var or an ensemble Kalman filter (Kalnay, 2003), we directly replace selected variables in the model state with corresponding truth fields from reanalysis during the autoregressive rollout. This idealized, noise-free intervention provides an upper-bound test of whether the model’s learned dynamics can propagate the inserted information to variables that are not directly modified, while avoiding uncertainties associated with observation operators, observation errors, background-error covariance specification, and assimilation-system design.

Using reanalysis gives our framework an observationally constrained reference. Unlike a free-running nature run, which may be evaluated against observations but does not assimilate them throughout the simulation, reanalysis incorporates real observations at each analysis time. Its fields remain model-based estimates, but they are anchored to observed atmospheric conditions.

Each model is integrated autoregressively for 60 days. During the first 30 days, the free-running forecast diverges from the reference and reaches error saturation. We then begin separate temperature, wind, and geopotential-height insertion experiments. By comparing each insertion experiment with a free-running control, we evaluate whether the models have learned physically consistent relationships involving hydrostatic balance, geostrophic balance, thermal-wind balance, moisture coupling, vertical structure, and land-ocean/terrain regime dependence, rather than assessing performance solely through reductions in forecast RMSE.

b. Scope and objectives

Using ERA5 and MERRA-2 as truth, each chosen to match the reanalysis distribution of model's own training data (Section 3), we test how responses vary with architecture, variable, latitude, surface regime, and terrain; the goal is to (i) provide a scalable ML-OSSE methodology and (ii) reveal the strengths, limitations, and physical readiness of AI weather models for assimilation-like observing-system design, while also contributing a direct-insertion ML-OSSE framework, a balance-based diagnostic of learned physical consistency, and a comparative analysis that pinpoints where these models can reliably complement or fall short of traditional NWP-based OSSEs.

## 2. Background and Related Work

OSSEs have guided atmospheric observing-system design since the early study by Charney et al. (1969). Using a two-level Mintz–Arakawa GCM, Charney et al. tested whether historical temperature observations could help recover the wind field. After generating a 170-day model history, they inserted reference temperatures during a 60-day integration and measured the resulting reduction in wind error. The inserted temperature information drove the simulated winds closer to the reference state, particularly in the extratropics, demonstrating that frequent and accurate temperature profiles can constrain both atmospheric winds and mass fields. Farhat et al. (2016) proved that, in the simplified 3D Planetary Geostrophic model, continuous nudging with coarse-mesh temperature observations can recover the reference temperature and velocity fields at an exponential rate. This provides a rigorous justification of the Charney conjecture for that model. Their result concerns nudging, which gradually adjusts the model toward observations; it does not establish stability for the hard-replacement method used here. Therefore, it does not rule out instabilities such as GraphCast's (Section 4.2.5).

The physical basis of this response lies in the coupling among temperature, geopotential, and wind. Hydrostatic balance links vertical temperature structure to geopotential thickness, while geostrophic and thermal-wind balance connect horizontal temperature and geopotential gradients to the wind field (Holton & Hakim, 2013). Temperature observations therefore provide not only direct temperature corrections but also indirect information about the broader atmospheric state, provided that the forecast system can propagate the information through these balance relationships.

Charney et al. (1969) reported a more nuanced result than a simple extratropical/tropical divide. Their own experiment found that wind could be recovered from temperature history at all latitudes, including the equator, where geostrophic balance does not hold – in their words, “the wind is determined at low latitudes,” though a much smaller temperature error (below about 0.25°C, versus values an order of magnitude larger at 49°N) was required for an appreciable reduction in wind error there. Ghil et al. (1977) later answered, for a barotropic primitive-equation system, the theoretical question Charney et al.’s experiment left open – whether the mass field and its time history determine the wind field exactly, not just approximately – deriving an exact diagnostic system whose validity is governed by an ellipticity condition on the flow’s divergence and deformation rather than on the Coriolis parameter alone, and is valid at all latitudes and regions for which the underlying primitive-equation model holds. Halem et al. (2000) revisited the original hypothesis using NASA’s higher-resolution GEOS GCM with improved physics and found a more restrictive result than Charney et al.’s: extratropical winds still adjusted strongly to temperature insertion, but tropical winds now required unrealistically small temperature errors to achieve comparable improvements – a divergence from, rather than a repetition of, the 1969 finding. These latitude-dependent outcomes continue to set accuracy and sampling requirements for modern infrared and microwave sounders.

NASA and NOAA researchers expanded OSSEs to evaluate proposed observing systems, including polar-orbiting sounders, satellite wind measurements, wind profilers, scatterometers, and Doppler wind lidar (Hoffman & Atlas, 2016). Atlas et al. (1985) found that simulated wind profiles reduced forecast errors more than temperature observations. Later OSSEs evaluated space-based Doppler wind lidar concepts (Rohaly & Krishnamurti, 1993; Atlas, 1997; Zhang & Pu, 2010), ERS and NSCAT scatterometer designs (Atlas, 1997), and infrared-sounder requirements for instruments such as AIRS. The Weather Research and Forecasting Innovation Act of 2017 formalized NOAA’s role in OSSEs: Section 107 directed the Assistant Administrator for Oceanic and Atmospheric Research to conduct OSSEs or other quantitative assessments of observing-system value and design options.

## 3. Methodology

We evaluate autoregressive AI weather models using a direct-insertion ML-OSSE design. Each model was advanced autoregressively in 6-hour steps. At each forecast time, we extracted the matching fields from the reference reanalysis and inserted them into the evolving model state. ERA5 (Hersbach et al., 2020) and MERRA-2 (Gelaro et al., 2017) provide the reference fields, which are temporally aligned with the forecasts and mapped to each model’s horizontal grid and vertical pressure levels. FourCastNet V2, Aurora, and GraphCast operate at ERA5’s native 0.25° resolution, so their ERA5 fields require no horizontal remapping. Similarly, Prithvi WxC operates at MERRA-2’s native 0.5° × 0.625° resolution, so its MERRA-2 fields require no horizontal remapping.

The primary truth source and model-specific processing are as follows:

FourCastNet V2 uses ERA5 at its native 0.25° grid resolution, with insertion applied through Earth2MIP (NVIDIA’s Earth-2 model-intercomparison and inference platform (NVIDIA, 2023)) at 13 pressure levels: 50, 100, 150, 200, 250, 300, 400, 500, 600, 700, 850, 925, and 1000 hPa, whereas the original ERA5 reanalysis resolves the atmosphere using 137 native hybrid sigma-pressure model levels and is also available at 37 standard pressure levels (NVIDIA, 2023).

Aurora: ERA5 is used as the primary truth source and mapped to the Aurora configuration used in this study (0.25°), 13 pressure levels: 50, 100, 150, 200, 250, 300, 400, 500, 600, 700, 850, 925, and 1000 hPa.

GraphCast uses ERA5 on its native 0.25° grid. Geopotential and wind are inserted at all 37 pressure levels. For temperature insertion, we exclude the 11 uppermost atmospheric levels, from 1 to 100 hPa, and insert temperature at the remaining 26 levels, from 125 to 1000 hPa. We exclude the uppermost levels because inserting temperature at all 37 levels causes severe degradation in GraphCast's geopotential and wind fields (Section 4.2.4).

Prithvi WxC uses MERRA-2 at its approximately 0.5° × 0.625° horizontal resolution, with insertion performed at the model's 14 selected pressure levels: 48, 109, 150, 208, 245, 288, 412, 525, 600, 700, 850, 925, 970, and 985 hPa, whereas the original MERRA-2 reanalysis provides 72 native hybrid sigma-pressure layers or 42 standard pressure levels. Because the nearest inserted level to the surface is 985 hPa, not exactly 1000 hPa, any figure reporting a '1000 hPa' result for Prithvi WxC reflects its nearest native output level rather than a directly-inserted one – unlike FourCastNet V2, Aurora, and GraphCast, which all insert truth at an exact 1000 hPa level.

Table 1 summarizes each model's exact checkpoint/variant, training target, and input configuration, since these details affect how insertion interacts with each model's own autoregressive mechanics.

| Model | Checkpoint / variant used | Training / prediction target | Input time steps per forecast step | Lead-time conditioning |
|---|---|---|---|---|
| FourCastNet V2 | Base SFNO checkpoint, autoregressively fine-tuned, via NVIDIA Earth2MIP | ERA5 | 1 (single most recent state) | Fixed 6-hour step |
| Aurora | Base pretrained checkpoint (not HRES-fine-tuned) | ERA5 | 2 (t−1, t) | Fixed 6-hour step |
| GraphCast | Standard 37-level operational checkpoint | ERA5 (1979–2017 training period) | 2 (t−1, t) | Fixed 6-hour step |
| Prithvi WxC | Rollout-fine-tuned checkpoint (prithvi.wxc.rollout.2300m.v1) | MERRA-2 | 2 (t−1, t) | Fixed 6-hour step (rollout-optimized variant; the base pretrained checkpoint also supports flexible/zero-hour lead-time conditioning, not used here) |

Cross-reanalysis experiments: CAMS fields are bilinearly interpolated using xESMF to the grid and pressure-level configuration of the corresponding model. These experiments test whether the response reflects learned physical coupling rather than merely familiarity with inserted data drawn from the same reanalysis distribution used during model training; this cross-reanalysis design was part of the study's methodology but its results are not carried through to the Results and Appendix sections reported in this version.

For each model, we run a 60-day autoregressive forecast with 6-hour steps. The model runs freely for the first 30 days. From day 30 onward, at each step, we replace the selected variable—temperature, wind, or geopotential height—with its corresponding ERA5 field for FourCastNet

V2, Aurora, and GraphCast, or MERRA-2 field for Prithvi WxC. We insert these fields at all pressure levels used in that experiment; no surface fields are inserted.

FourCastNet V2 uses one input frame, so the corrected output becomes the input to the next forecast step. Aurora, GraphCast, and Prithvi WxC use two input frames. At the first insertion, the newest frame is corrected while the older frame remains from the free run. At the next 6-hour step, both frames are corrected. Each subsequent step adds one newly corrected frame as the previous corrected frame shifts into the older input slot.

Additional experiments include geopotential-height insertion and the insertion of combined thermodynamic–dynamic states. After each insertion, the forecast proceeds autoregressively, requiring the model to propagate the inserted information to non-inserted variables and subsequent forecast steps through its learned dynamics. The 60-day horizon balances two constraints: it is long enough to establish a stable, multi-week mature-phase average (days 30-60) for computing RMSE and coupling scores, consistent with Charney et al.'s (1969) own 60-day post-insertion integration window, while remaining short enough to keep the cost of running this design across four architectures and multiple insertion experiments computationally tractable; Section 7 discusses how a longer or ensemble-based extension could further test the robustness of these conclusions. In practice, each 60-day (240-step) experiment completed in roughly 30 minutes to 2 hours of wall-clock time, depending on the model and the GPU used, which makes this AI-based OSSE framework fast and inexpensive compared with NWP-based OSSEs that require full model integrations and data-assimilation cycles.

Extending GraphCast inference to 60 days

GraphCast's public inference pipeline (Lam et al., 2023) is built around fixed example datasets: the target template and the forcing fields for every forecast step are read directly from the example file, which covers at most 10 days (40 six-hour steps). A 60-day (240-step) rollout therefore required us to rebuild the inference setup manually (Diggikar et al., 2026c). First, we assembled a new 1 February 2023 initial-condition file from ERA5 at all 37 pressure levels in the exact schema of DeepMind's example data, with the two input times (00 and 06 UTC), the static fields (surface geopotential and land–sea mask) taken from the example file, and 6-hour accumulated precipitation computed from hourly ERA5 forecast fields. Second, we replaced the fixed target template with a 240-step template of lazily evaluated placeholder arrays, which defines the shape and valid time of every future step without holding them in memory. Third, because GraphCast's forcings (top-of-atmosphere incident solar radiation and the sine and cosine of the year and day progress) depend only on date, time, and location, we recomputed them from astronomical formulas for each of the 240 future valid times instead of reading them from data. Finally, the rollout runs in chunks of four steps, each written to disk and then released, so memory use does not grow with forecast length; the model integrates all 37 levels, while 13 levels are saved for analysis to match the other models. In the insertion experiments, a lead-time-indexed ERA5 truth file covering February–April 2023 supplies the inserted fields, which overwrite the model's predictions before they are fed back as inputs for the next step.

Model responses are evaluated using root-mean-square error (RMSE) and spatial bias, following the latitude-weighted RMSE evaluation convention established for data-driven weather forecasting (Rasp et al., 2020). These diagnostics are calculated for both directly inserted variables and the secondary responses of non-inserted atmospheric fields. The effects of global insertion are further stratified by latitude band, cloud regime, and terrain characteristics to determine how physical coupling varies between the tropics and extratropics and across different atmospheric and surface conditions.

Extended experiments examine the sensitivity of the response to the spatial coverage and accuracy of the inserted observations. Using land–sea masks derived for the corresponding datasets, insertion is performed separately at land-only and ocean-only grid coordinates. Perturbed-insertion experiments are also conducted by adding controlled random errors to the global truth fields before insertion. Together, these experiments assess how model responses depend on observation location, surface regime, and observational uncertainty, revealing where and under which insertion configurations AI weather models can most effectively complement traditional OSSE practice.

Perturbation generation (repeated every six hours): The COP field is regenerated from scratch at every insertion time using the following procedure, applied independently on each native pressure level:

1. Base random field: Draw standard-normal white noise $n_\ell(\varphi, \lambda) \sim N(0, 1)$ where $\ell$ indexes levels, and $(\varphi, \lambda)$ are latitude–longitude grid points.

Horizontal correlation : Apply an isotropic Gaussian filter that yields an e-folding horizontal correlation length of 250 km, consistent with satellite radiance observation-error correlation length scales reported for infrared sounders (Bormann et al., 2016).

- On 0.25° × 0.25° grids (FourCastNet V2, Aurora) this corresponds to a kernel half-width of 9 grid cells.
- On Prithvi WxC's 0.5° × 0.625° grid the half-widths are 4.5 cells meridionally and 3.6 cells zonally. (True great-circle isotropy degrades poleward, a limitation)

2. Vertical correlation: At each grid point $(\varphi, \lambda)$, stack the smoothed fields across levels into $s(\varphi, \lambda) = [s_1, \ldots, s_L]^T$, then apply the Cholesky factor $A$ of the target vertical-correlation matrix:

$$m(\varphi, \lambda) = A\, s(\varphi, \lambda).$$

Here, $m_\ell$ is the vertically correlated perturbation at level $\ell$, before amplitude rescaling.

3. Amplitude rescaling: Each level $\ell$ of m is rescaled to its target physical standard deviation, following an Ingleby-style error-by-pressure table (Ingleby, 2017).

- Temperature (50–1000 hPa): 2.0, 1.5, 1.5, 1.2, 1.0, 0.8, 0.7, 0.7, 0.8, 0.8, 1.0, 1.0, 1.2 K.
- Wind (same levels): 1.5, 1.6, 1.8, 2.2, 2.0, 1.8, 1.4, 1.2, 1.1, 1.0, 1.0, 1.1, 1.3 m $s^{-1}$.

4. Non-divergent wind (wind only): Instead of perturbing u and v directly, draw a scalar stream-function perturbation $\psi'_\ell$ using Steps 1–4, then compute

$$u' = -(1/R)\, \partial\psi'/\partial\varphi,$$

$$v' = (1 / (R \cos \varphi))\, \partial\psi'/\partial\lambda,$$

clipping cos φ below cos 85° to avoid the polar singularity. Finally rescale u′, v′ to the wind σ-table above (assuming $\sigma_u = \sigma_v$).

5. State replacement: Add the perturbation to the verifying analysis and hard-insert the sum into the model state every 6 h from day 30 to day 60. Two variants of this design are evaluated: a temperature-only perturbation, inserted at all pressure levels, and a combined zonal-and-meridional-wind-only perturbation, also inserted at all pressure levels; both follow this same day-30-to-60, 6-hourly hard-insertion protocol.

## 4. Results & Analysis

We compare a free-running control with direct-insertion experiments in which selected atmospheric variables including temperature, wind, geopotential height, and combined state variables are overwritten every 6 h from day 30 onward using the corresponding truth fields. Forecast skill is evaluated using RMSE and spatial bias at representative tropospheric levels for temperature, winds, and geopotential height. Three periods are distinguished: (1) pre-insertion (days 0–30, when the control and insertion runs are identical), (2) adjustment (days 31–40, capturing the immediate model response), and (3) mature response (days 40–60, the stabilized impact window used for diagnostics). Results are evaluated globally and stratified across latitude bands, cloud regimes, terrain classes, and land-ocean regions to identify how information propagation varies across atmospheric and surface conditions.

### 4.1 Control Run

All four architectures were first run freely for 60 days (240 six-hour steps) to establish baseline error growth (see the Temperature RMSE curves in Fig. 1 and Zonal Wind RMSE in Fig. 2):

• FourCastNet V2: initialized from ERA5 at 00 UTC 1 Feb 2023; ended 12 UTC 1 Apr 2023.

• Aurora: two-step start (00 UTC & 06 UTC 1 Feb 2023); ended 00 UTC 2 Apr 2023.

• GraphCast: two-step start (00 UTC & 06 UTC 1 Feb 2023); ended 06 UTC 2 Apr 2023.

• Prithvi WxC: two-step start (00 UTC & 06 UTC 1 Feb 2023); ended 06 UTC 2 Apr 2023.

All four models share the same initialization window (1 Feb 2023), which removes a potential confound in which cross-model comparisons could otherwise partly reflect differences in the specific 60-day weather situation sampled by each model, rather than purely architectural differences.

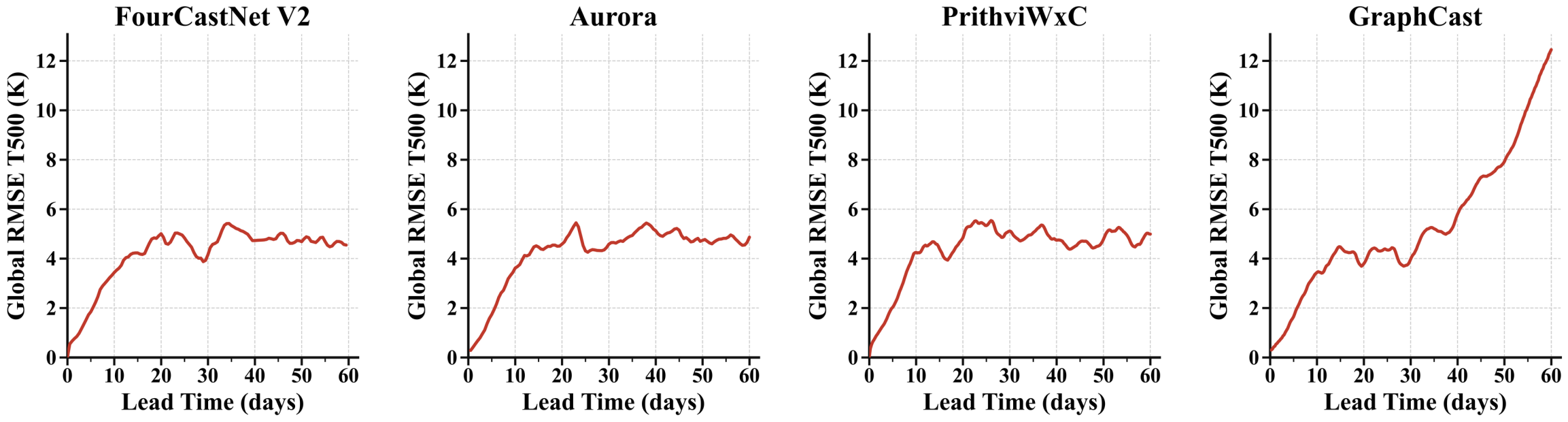

**Fig. 1.** Global cosine-latitude-weighted Temperature RMSE at 500 hPa for free-running control experiments over a 60-day lead time. Results are shown for FourCastNet V2, Aurora, PrithviWxC, and GraphCast with no observation insertion applied.

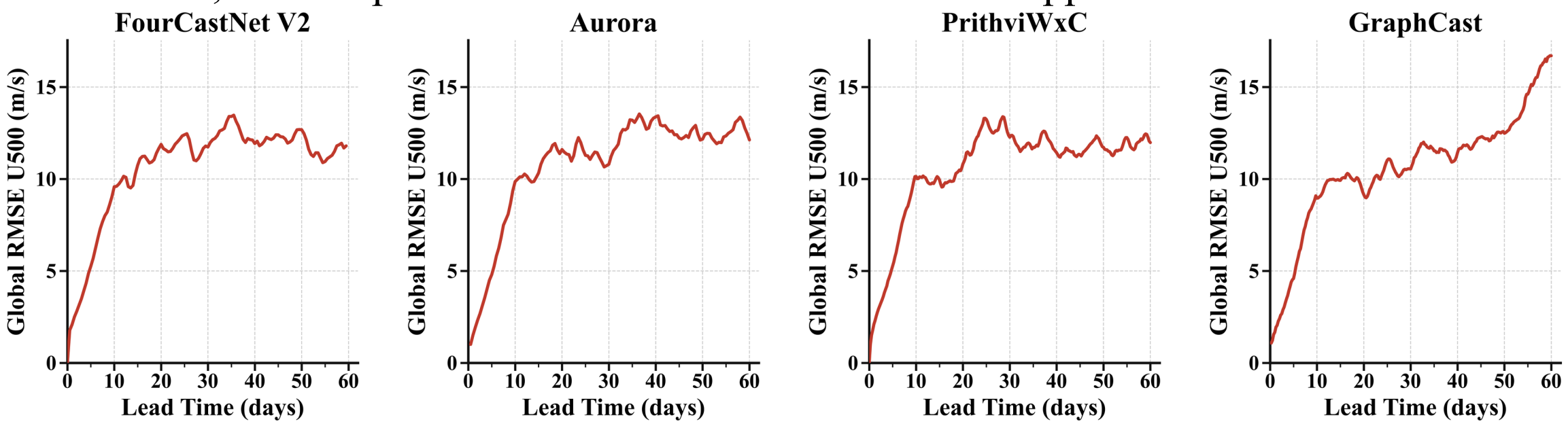

**Fig. 2.** Global Zonal Wind RMSE at 500 hPa for free-running control experiments over a 60-day lead time. Results are shown for FourCastNet V2, Aurora, PrithviWxC, and GraphCast with no observation insertion applied.

## 4.2 Temperature Insertion

After a 30-day free-running spin-up, we overwrite every pressure-level temperature with reanalysis truth at 6-hour intervals for the remaining 30 days, leaving all other fields untouched. The inserted temperature RMSE is calculated immediately after each 6-hourly insertion, during days 30–60. Because the model temperature is overwritten with reanalysis truth before RMSE is calculated, the RMSE is zero by construction, as shown in Figure 3. This confirms that the overwrite is applied as intended; it does not measure the model's temperature forecast skill.

This creates an idealized full-truth thermodynamic constraint while leaving the remaining variables, including wind, geopotential, moisture, and surface fields, free to evolve through the model's learned dynamics.

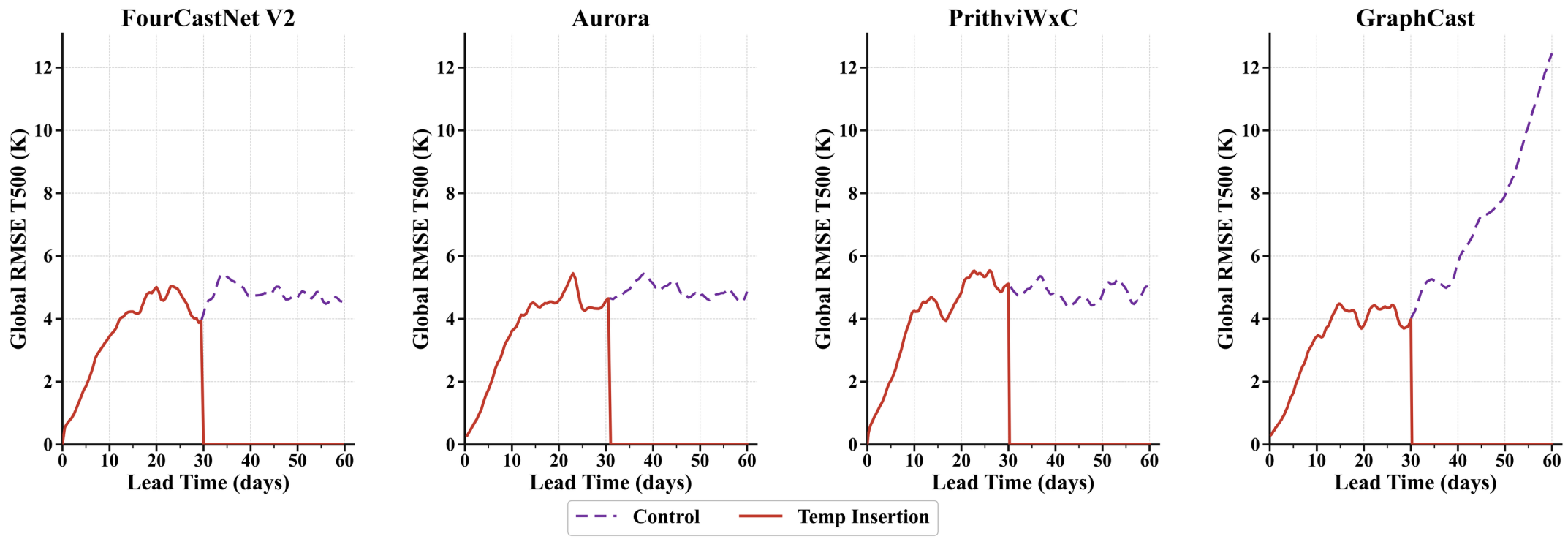

**Fig. 3.** Global RMSE of temperature at 500 hPa (T500) for the temperature insertion experiment across all four AI weather models.

### 4.2.1 Experimental Question

Does accurate temperature forcing alone enable the model to regenerate dynamically consistent mass and wind fields? In Charney's conjecture (Charney et al., 1969), success would imply that the network has internalized hydrostatic links between temperature and geopotential thickness and thermal-wind links between horizontal temperature gradients and vertical wind shear (Holton & Hakim, 2013). Because these balances are learned, not hard-wired, any improvement in non-inserted variables (geopotential height, winds) after temperature insertion reveals genuine physical coupling within the model; confinement of the benefit to temperature alone signals missing or weakly learned dynamics.

#### 4.2.2 Impact of Temperature Insertion on Zonal Winds

The key physical pathway in this experiment is not a direct temperature-to-wind replacement, but a sequence of balance adjustments that the model must reproduce internally. When temperature is overwritten with the reference state, the model receives a corrected thermal structure. In the real atmosphere, this corrected thermal structure modifies layer thickness through hydrostatic balance: warmer columns correspond to larger geopotential thickness, while colder columns correspond to smaller thickness. This produces changes in the geopotential height field. Once geopotential gradients are modified, the horizontal pressure-gradient force changes, and the balanced wind field should adjust through geostrophic and thermal-wind relationships. Therefore, a successful temperature-to-wind response requires the model to connect four steps: temperature correction, vertical thickness correction, horizontal geopotential-gradient correction, and finally wind adjustment. In classical OSSE terms, this is closely related to the Charney-style question of whether temperature information can determine the large-scale wind field. However, in the AI-model setting, there is no explicit hydrostatic solver, pressure-gradient calculation, or geostrophic adjustment operator. Temperature insertion separates the four models into two broad groups: a group that recovers zonal wind strongly and consistently across latitude (Aurora, FourCastNet V2, and Prithvi WxC), and a group whose recovery is weaker and more latitude-dependent (GraphCast) (Fig. 4, columns left to right).

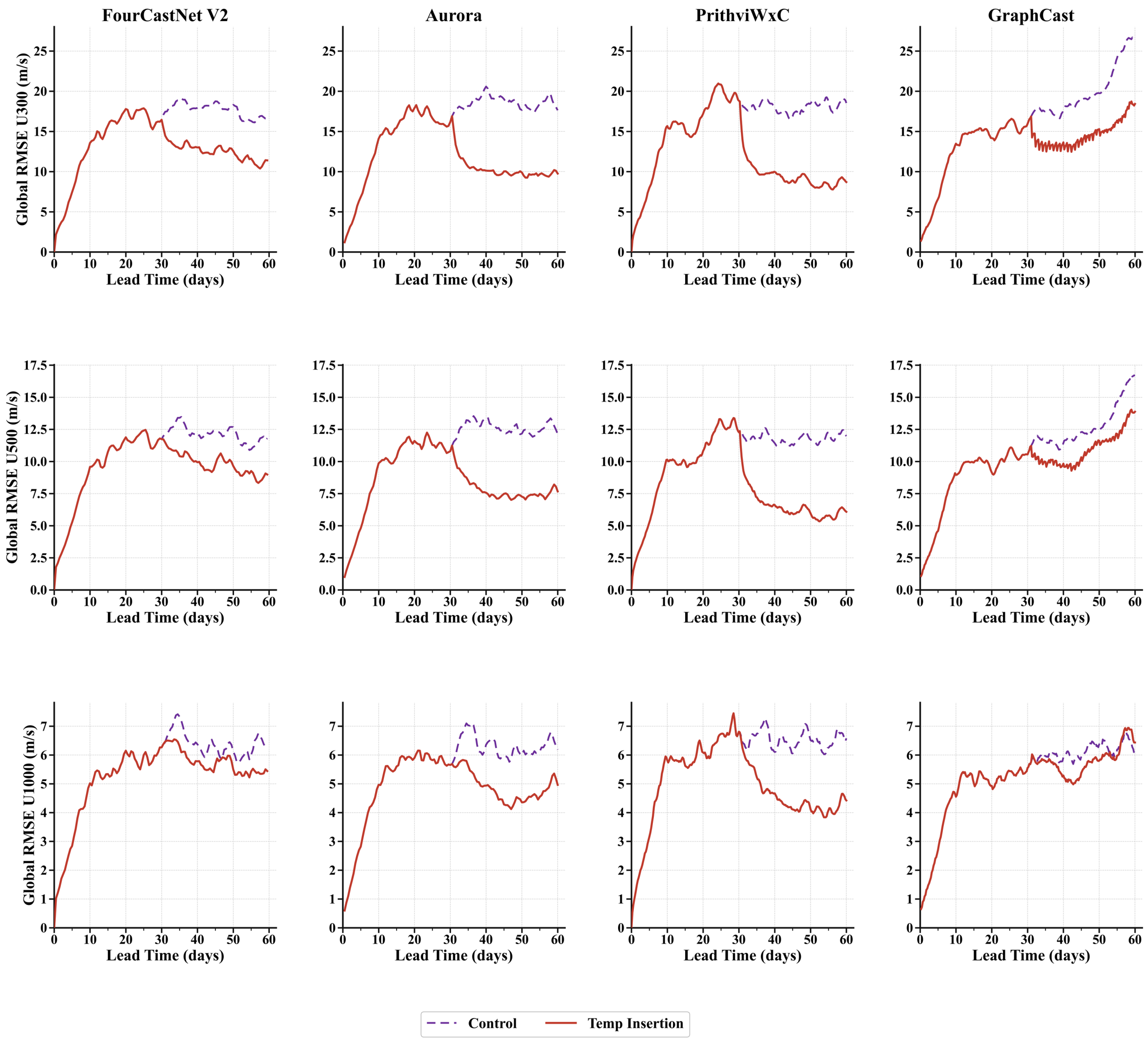

**Fig. 4.** Global RMSE of zonal wind (U) at 300, 500, and 1000 hPa following temperature insertion across all four AI weather models.

FourCastNet V2 and Aurora display sustained RMSE drops in zonal wind from day 30 onward. Prithvi WxC shows a comparably sustained drop: its mature-phase (day 30-60) U-wind coupling score, averaged over all pressure levels, is 42.1%, close to Aurora's 38.6% and well above FourCastNet V2's 18.0% on this particular pathway (Appendix B3, Fig. B4). GraphCast's gains remain the weakest and most latitude-dependent of the four, and are negative at the two lowest latitude bands examined (see Section 4.2.3). This response is physically consistent with atmospheric dynamics. Hydrostatic and thermal-wind balance link temperature to geopotential and wind. It is also qualitatively similar to classical physics-based OSSE results: Charney et al. (1969) [14] used a circulation model governed by physical equations and found that repeatedly inserting temperature information improved the inferred wind field. We observe a similar response in several AI weather models, although its strength varies across models and latitudes. A plausible architectural explanation for this spread follows. Aurora succeeds because its shared 3-D latent state (Bodnar et al., 2025) is multivariate and vertically coherent – i.e., it represents multiple pressure levels jointly in one shared latent space rather than encoding each level independently – so corrected temperature immediately informs wind and height. FourCastNet V2 benefits from global spectral mixing (Pathak et al., 2022; Bonev et al., 2023; Li et al., 2021; Guibas et al., 2022), which projects thermal increments onto smooth synoptic momentum fields. Prithvi WxC's 2-D token geometry and reconstruction-heavy objective (Schmude et al., 2024; Vaswani et al., 2017; Dosovitskiy et al., 2021) evidently allow a substantial temperature-to-wind pathway to be learned, even though its representation still favors local column repair in some respects (see the persistent wind-to-geopotential contrast discussed below). GraphCast's temperature-to-wind response is partial and geographically uneven. Its multimesh combines fine-scale local edges with coarse-scale edges from all icosahedral refinement levels that bridge long distances. Its processor applies 16 unshared message-passing layers across this structure (Lam et al., 2023) [2]. This approach builds on GNN message-passing frameworks formalized by Gilmer et al. (2017) [10] and Battaglia et al. (2018) [11], and applied to weather forecasting by Keisler (2022) [30] and Lam et al. (2023) [2]. The cause of GraphCast's weaker response remains unclear. In our experiments, temperature insertion worsens wind RMSE in the two lowest latitude bands, whereas the other three architectures show improvement.

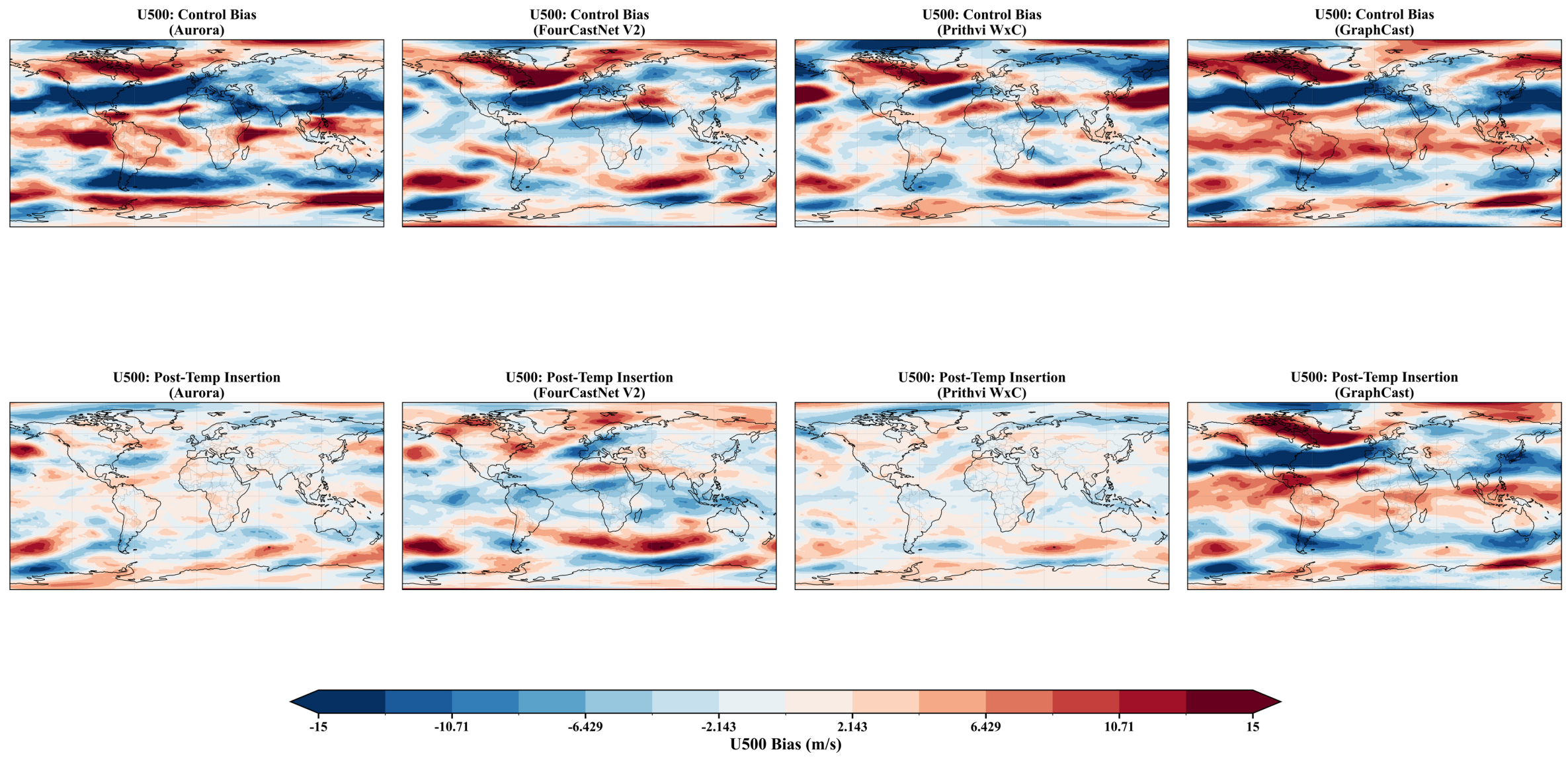

**Fig. 5.** Spatial bias in U500 (m/s) during the assimilation phase (days 30-60): each model's own free-running control bias (top row) versus its own post-temperature-insertion bias (bottom row), for Aurora, FourCastNet V2, Prithvi WxC, and GraphCast (left to right). Positive (red) and negative (blue) values indicate where the model overestimates or underestimates ERA5 zonal wind. Shared discrete color bar; values beyond ±15 m/s are clipped.

Fig. 5 shows this directly for U500: the post-insertion bias field is visibly flatter than the control-run field for Aurora, FourCastNet V2, and Prithvi WxC, while GraphCast's pattern changes comparatively little.

#### 4.2.3 Latitudinal response to global temperature insertion

Global temperature overwrites (days 30-60) expose clear latitude-dependent patterns in each model's learned balance (see Fig. 6).

Hydrostatic pathway (T → Z) dominates for three of the four models: Aurora and FourCastNet V2 improve geopotential-height RMSE by ≈ 66-82 % and 19-40 % respectively across the four latitude bands sampled (2N, 14N, 40N, 60N); Prithvi WxC improves ≈ 41 % in the deep tropics rising to ≈ 63-68 % from the mid-latitudes to the pole, a broad and consistent hydrostatic response, not a polar-only one. GraphCast, by contrast, shows a strongly adverse Z500 response at 2N (-53.0%) and even more strongly adverse at 14N (-161.9%), turning modestly positive at 40N (+5.3%), and adverse again at 60N (-35.1%). This latitudinal pattern for GraphCast's temperature-to-geopotential pathway is consistent with the control-run RMSE growth after day 40 noted in Section 4.1.

Geostrophic/thermal-wind pathway (T → U/V) is latitude-dependent for every model, though not always in the direction classical Coriolis scaling predicts. FourCastNet V2 is weakest in the deep tropics as expected on U500 (≈6% at 2N, rising to ≈23-35% poleward), and GraphCast's impact is clearly adverse at the two lower-latitude bands (-8.5% at 2N, -11.0% at 14N) before turning positive poleward (≈15% at 40N and 60N), implying geostrophy is learned mainly poleward of roughly 30° for this model. Prithvi WxC is also weakest in the deep tropics (≈31% at 2N) but otherwise strong and fairly flat across the remaining three bands (≈44-52%). Aurora, by contrast, does not taper toward the tropics at all on U500: its response is strongest at 2N (≈52%) and weakest at 60N (≈33%) – the opposite latitudinal pattern from FourCastNet V2, Prithvi WxC, and classical expectation. Despite this difference in latitudinal shape, the two foundation-model architectures retain the most skill specifically in the deep tropics: Aurora's and Prithvi WxC's U500 values at 2N (≈52% and ≈31% RMSE improvement) remain well above FourCastNet V2's ≈6% and far above GraphCast's outright negative value there, indicating that the two foundation-model architectures retain meaningful tropical temperature-to-wind skill where classical balance arguments alone would predict little benefit.

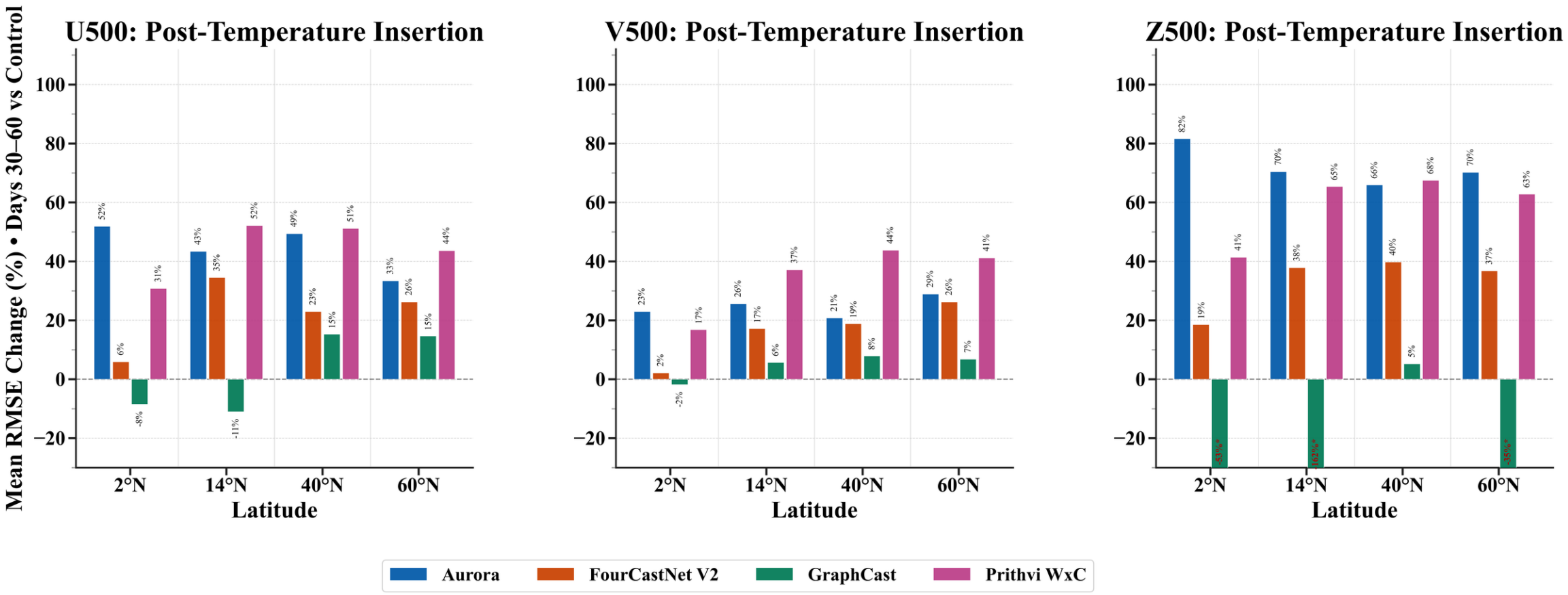

**Fig. 6.** Cross-variable RMSE response to temperature insertion during the assimilation phase (lead days 30–60). Panels show the mean percentage reduction in RMSE relative to the free-running control for zonal wind (U), meridional wind (V), and geopotential height (Z) at 500 hPa, evaluated at latitudes 2°, 16°, 40°, and 60°. Values are averaged over days 30–60.

4.2.4 Impact of Temperature Insertion on Geopotential

The temperature-to-geopotential link is much stronger than the temperature-to-wind link for three of the four models: after day 30, FourCastNet V2, Aurora, and Prithvi WxC sharply reduce mid-tropospheric geopotential-height RMSE across latitude. GraphCast is the exception its temperature-to-geopotential response is adverse across most of the sampled latitude range, even though its temperature-to-wind response is mildly positive (Fig. 7; see Section 4.2.3 for the latitude-resolved values).

When temperature is overwritten with the reference state, the model receives a corrected vertical thermal structure. In the real atmosphere, hydrostatic balance directly links this temperature structure to the geopotential thickness between pressure levels (Holton & Hakim, 2013): warmer layers occupy greater vertical thickness, whereas colder layers occupy less. The corrected temperatures should therefore modify layer thicknesses, which accumulate vertically to adjust geopotential height at successive pressure levels. A successful temperature-to-geopotential response thus requires the model to connect three steps: temperature correction, hydrostatic thickness adjustment, and vertical integration of those thickness changes into the geopotential-height field. Because this pathway is more direct than the additional pressure-gradient and momentum adjustments required for a temperature-to-wind response, temperature insertion is generally expected to influence geopotential more strongly and consistently than wind, and this holds true for Aurora, FourCastNet V2, and Prithvi WxC.

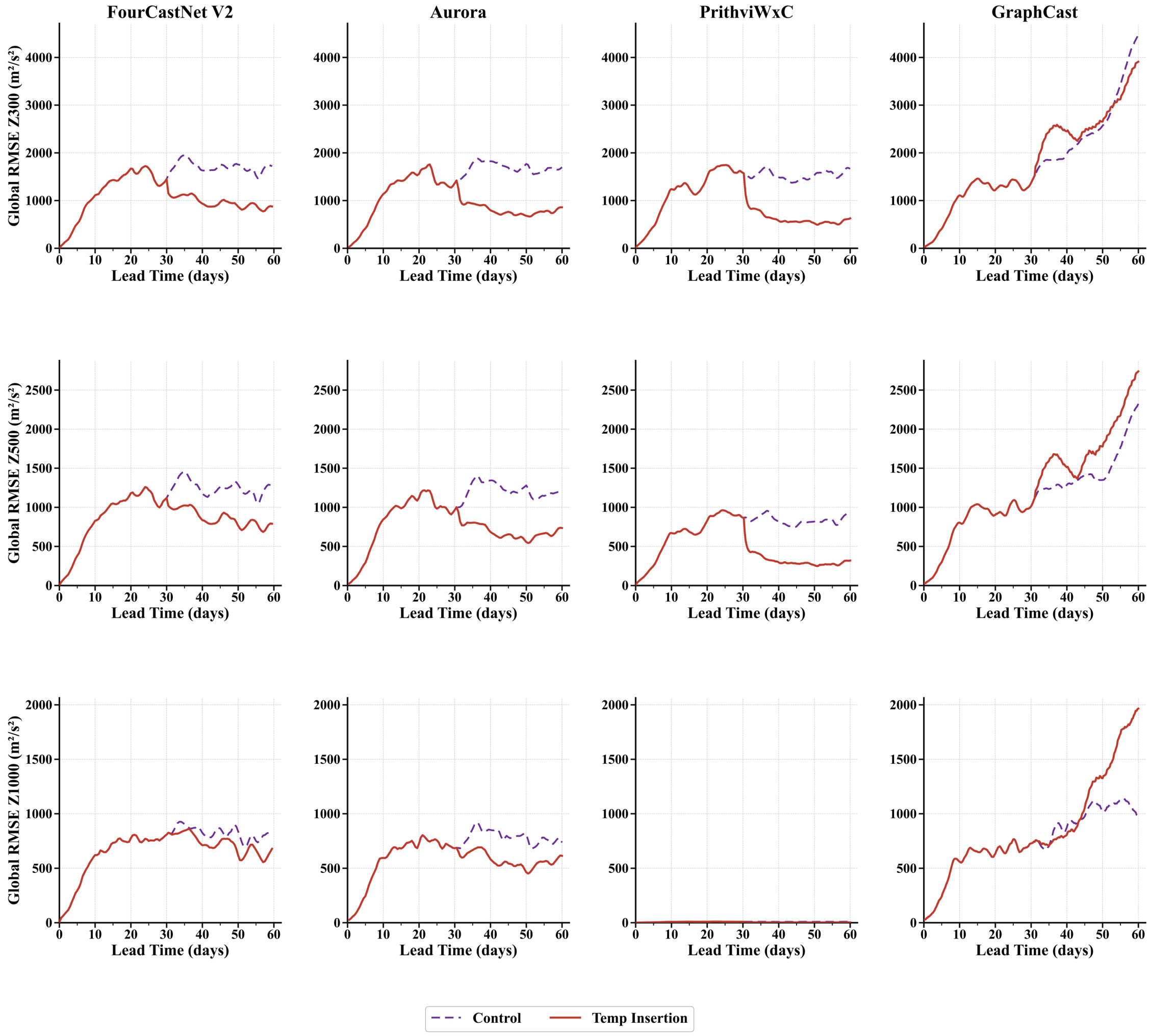

**Fig. 7.** Global RMSE of Geopotential (Z) at 300, 500, and 1000 hPa following temperature insertion across all four AI weather models.

The asymmetry between the temperature-to-geopotential and temperature-to-wind pathways is expected on physical grounds: once temperature is fixed, recovering geopotential merely requires recalculating column mass, which is generally easier than reconstructing momentum. Prithvi WxC does well here, consistent with a representation (Schmude et al., 2024) that captures vertical thermo-mass covariances; its temperature-to-wind coupling (Section 4.2.2-4.2.3) is comparable to Aurora's, so this strength should not be framed as happening "despite weak horizontal balance." GraphCast lags on this specific pathway: temperature is evidently too indirect a path to geopotential height for this model, and its graph-based architecture (Lam et al., 2023) appears to assimilate geopotential-height corrections directly far more effectively than it can infer them from thermal overwrites alone. Consequently, GraphCast improves far more when geopotential itself is inserted than when temperature is used as its proxy, since its temperature-to-geopotential pathway is negative at most latitudes.

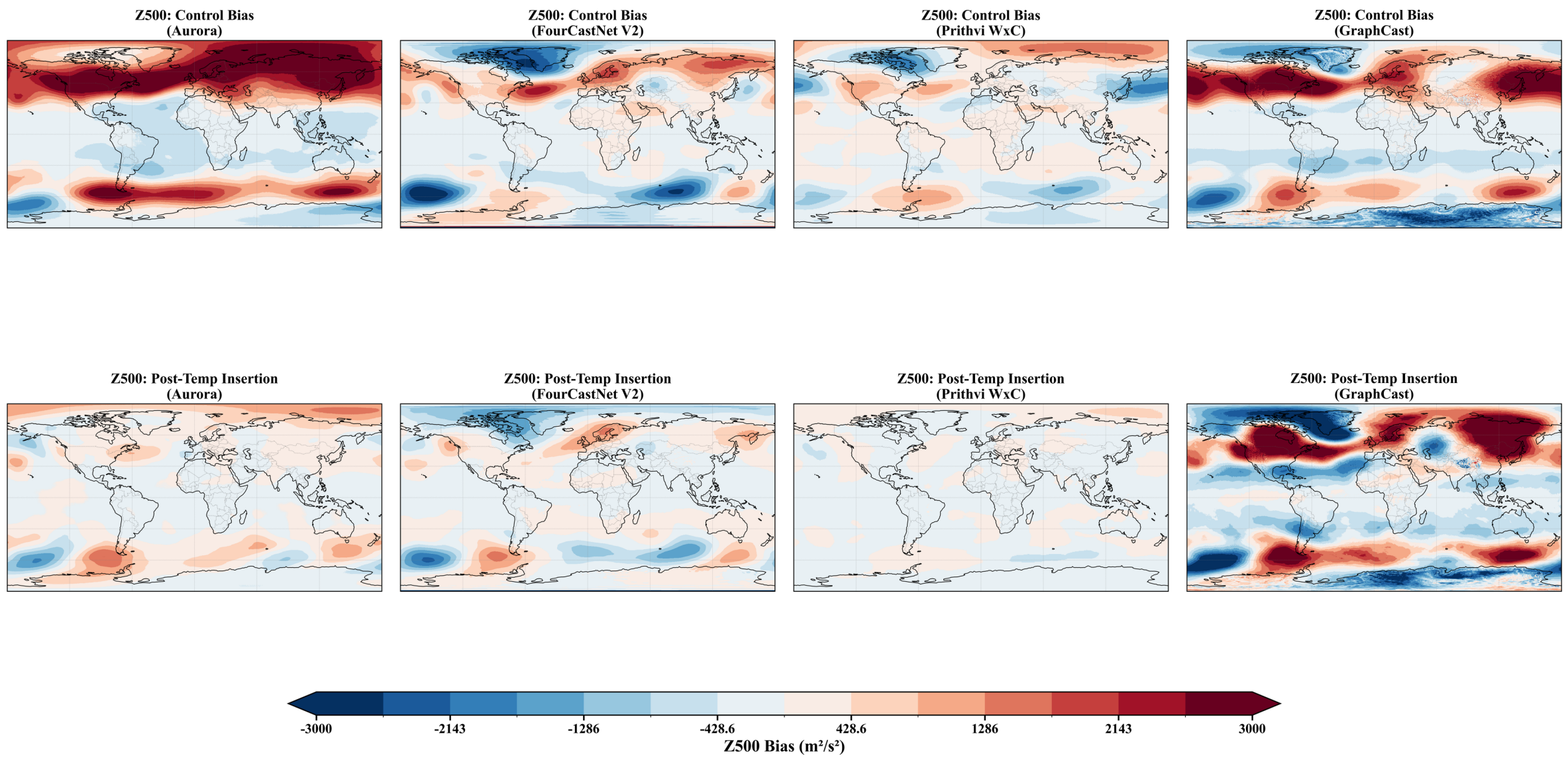

**Fig. 8.** Spatial bias in Z500 ($m^2/s^2$) during the assimilation phase (days 30-60): each model's own free-running control bias (top row) versus its own post-temperature-insertion bias (bottom row), for Aurora, FourCastNet V2, Prithvi WxC, and GraphCast (left to right).

Fig. 8 shows the same flattening for Z500: Aurora, FourCastNet V2, and Prithvi WxC's post-insertion bias fields are visibly reduced relative to their control-run patterns, while GraphCast's remains as pronounced as its control bias, consistent with its adverse temperature-to-geopotential coupling quantified above.

#### 4.2.5 GraphCast-Specific Anomaly Under Temperature Insertion

GraphCast's full native configuration operates at 37 pressure levels. Using the same 2023-02-01 initial condition as the rest of this study, a dedicated experiment inserted temperature at this full 37-level configuration to characterize GraphCast's behavior under maximal vertical forcing and to identify a practical mitigation for any failure found.

With all 37 pressure levels hard-replaced by ERA5 truth every 6 hours from day 30 onward, temperature insertion produces a severe GraphCast failure. Geopotential insertion and wind insertion at the same 37-level configuration do not show this behavior and improve the other fields. By day 60, geopotential reaches approximately 7720 model units, compared with 1748 in the free-running control, while zonal wind reaches -1894.6 m/s at 50 hPa near the South Pole. The mature-phase mean coupling scores are -195.7% for zonal wind (U), - 212.6% for meridional wind (V), and -199.6% for geopotential (Z). The degradation is concentrated at the uppermost, lowest-pressure levels.

A normalized shock analysis comparing the control run's error with the natural variability for the same variable, level, and region dentifies temperature at 50 hPa near the South Pole as the largest anomaly among the combinations tested (shock z-score 1.86; the other combinations are roughly 0.9). This location also coincides with a weakness in GraphCast's grid-to-mesh mapping: at the pole, all longitude points map to a single coordinate. These results suggest that repeatedly inserting temperature here may contribute to the failure, but they do not establish the cause. One possibility specific to this GraphCast experiment is that the model does not sufficiently damp the local perturbation between 6-hourly insertions, allowing the geopotential and wind errors to grow.

As a direct test of this explanation, the same experiment was repeated with the top 11 pressure levels (all levels at or below 100 hPa) excluded from insertion, leaving the remaining 26 levels (125 hPa through 1000 hPa) hard-replaced as before. The rescue is dramatic: zonal wind flips to positive at every saved level, from +11.1% (50 hPa) to +34.9% (150 hPa) improvement over control; meridional wind is likewise positive at every level, from +3.8% (700 hPa) to +14.9% (300 hPa); geopotential is mixed but far more contained than the uncontrolled blow-up seen with all 37 levels inserted - clearly positive at 850 hPa (+23.4%), 925 hPa (+12.6%), 700 hPa (+5.1%), and 200 hPa (+2.3%), with the remaining levels still negative but bounded, ranging from -2.9% (600 hPa) to -63.6% (100 hPa) - nothing resembling the several-hundred-percent uncontrolled divergence seen when all 37 levels are inserted. Temperature itself still improves by approximately 99% at the levels that remain inserted (e.g., 99.17-99.23% at 150-300 hPa), confirming that withholding the vulnerable upper levels does not meaningfully cost the correction its benefit to the temperature field itself. Wind insertion and geopotential insertion are unaffected by this issue and continue to use GraphCast's full 37-level configuration throughout this study, since only temperature insertion produces the runaway feedback described here.

### 4.3 Wind Insertion

After a 30-day free-running spin-up, we overwrite every pressure-level zonal and meridional wind with reanalysis truth at 6-hour intervals for the remaining 30 days, leaving all other fields untouched. All models track the injected winds almost perfectly from day 30 onward, confirming that the overwrite works as intended (see Fig. 9).

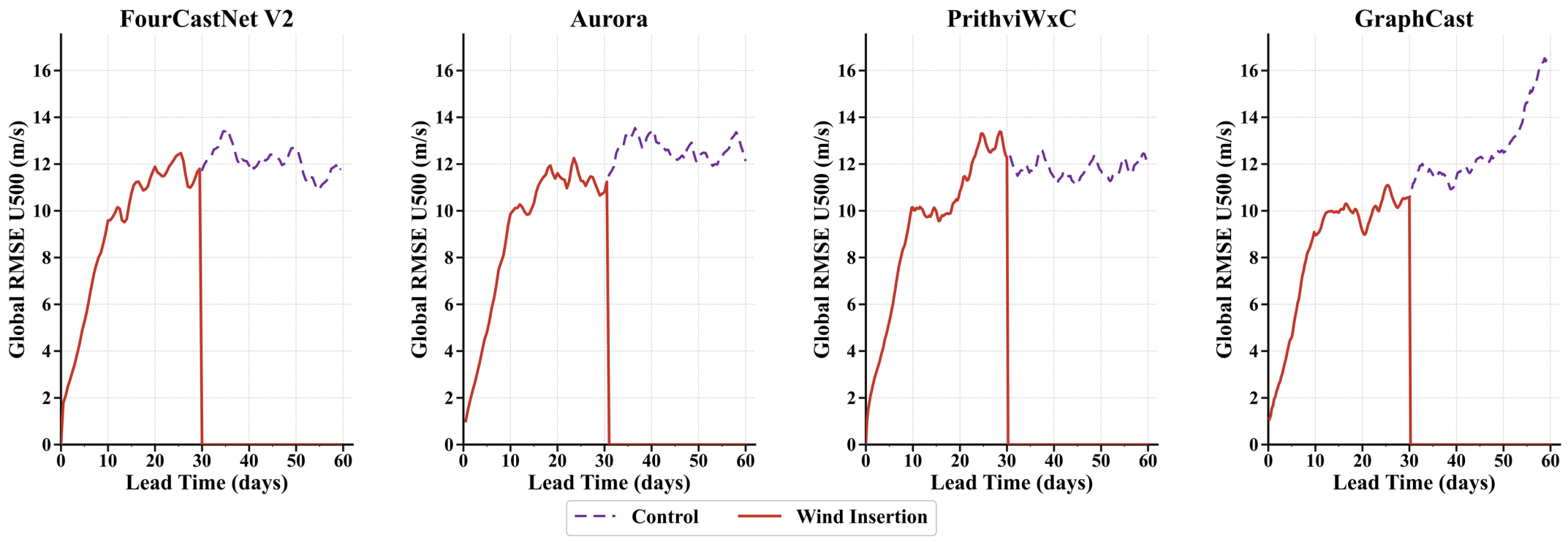

**Fig. 9.** Global RMSE of zonal wind at 500 hPa (u500) for the wind insertion experiment across all four AI weather models.

#### 4.3.1 Experimental Question

Wind insertion asks a complementary question: if the rotational flow is corrected after the forecast has diverged, can the model recover the thermal and mass fields without directly overwriting them? This is a cleaner test of large-scale dynamical control than temperature insertion, because wind errors project directly onto advection, vorticity evolution, and wave propagation. In OSSE terms, it asks whether a forecast model can use wind observations as broad state constraints rather than as corrections confined to the wind channel alone.

#### 4.3.2 Impact of Wind Insertion on Temperature

When the wind field is overwritten with the reference state, the model receives a corrected atmospheric circulation. Winds influence temperature primarily by transporting warm and cold air through horizontal advection. Corrected winds also modify convergence and divergence, which affect vertical motion: rising air cools through adiabatic expansion, whereas descending air warms through compression. Changes in near-surface circulation can additionally alter surface-atmosphere heat and moisture exchange. A successful wind-to-temperature response therefore requires the model to connect four processes: wind correction, horizontal heat transport, circulation-induced vertical motion, and the resulting thermal adjustment (Holton & Hakim, 2013). This coupling is generally stronger in the extratropics, where winds and temperature gradients are closely linked through geostrophic and thermal-wind balance but may be weaker or less stable in the tropics, where divergent flow, convection, and moisture processes play a greater role.

Wind insertion yields a strong secondary temperature response in Aurora, FourCastNet V2, and Prithvi WxC, while GraphCast's response remains weak and turns adverse at low latitudes (Fig. 10). Numerically, the mature-phase (day 30-60), all-pressure-level-averaged coupling summary (Appendix B3, Fig. B4) gives wind-to-temperature scores of 71.3% for Prithvi WxC, 68.9% for Aurora, 55.4% for FourCastNet V2, and 27.7% for GraphCast: Prithvi WxC's wind-to-temperature coupling score is in fact the highest of the four models on this pathway. FourCastNet shows broad positive wind-to-temperature behavior in the latitudinal and spatial analyses. In the midlatitudes, Aurora’s T500 improvement rises from roughly 54% in the tropics to around 74–88% poleward of the tropical belt, while FourCastNet V2 increases from roughly 32% near the equator to around 71% near 40°N. Prithvi WxC's T500 response shows the same tropics-to-extratropics tapering: 24.5% at 2N, rising to 52.6% at 14N and to 82.7-81.3% at 40N and 60N respectively - comparable in shape and, at the two higher-latitude bands, comparable in magnitude to Aurora's own response.

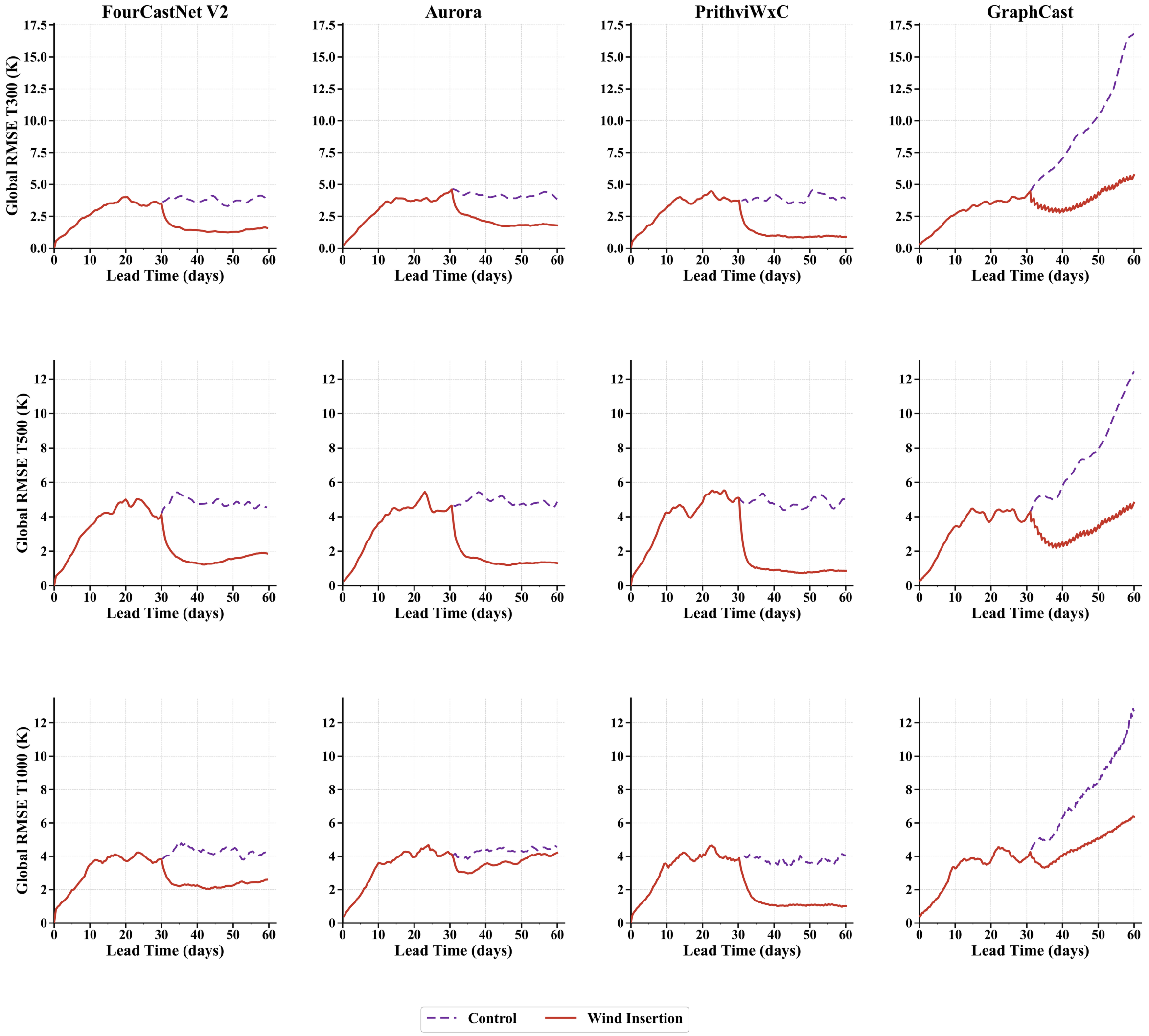

**Fig. 10.** Global RMSE of Temperature (T) at 300, 500, and 1000 hPa following wind insertion.

This ranking follows directly from how each model organizes circulation information. Aurora's multivariate 3D latent state (Bodnar et al., 2025) and roll-out fine-tuning make wind a controlling variable for the larger state. FourCastNet V2's global operator (Pathak et al., 2022; Bonev et al., 2023; Li et al., 2021; Guibas et al., 2022) likewise turns wind corrections into large-scale thermal improvement because the spectral representation directly tracks the synoptic wave structure that carries temperature advection. Prithvi WxC's representation (Schmude et al., 2024) evidently also translates an isolated wind correction into thermal rebalance effectively, and does so more strongly, on this composite metric, than any of the other three models. GraphCast (Lam et al., 2023) improves temperature only where corrected winds align with the learned extratropical circulation machinery: wind insertion worsens T500 error at the two lowest latitude bands (-41.3% at 2N, -10.9% at 14N) before turning strongly positive at 40N and 60N (+69.0%, +61.2%), which indicates that GraphCast has learned a largely geostrophic-style balance scheme that does not generalize to divergent tropical flow.

Spatial T500 bias (Fig. 11.) before and after wind insertion.

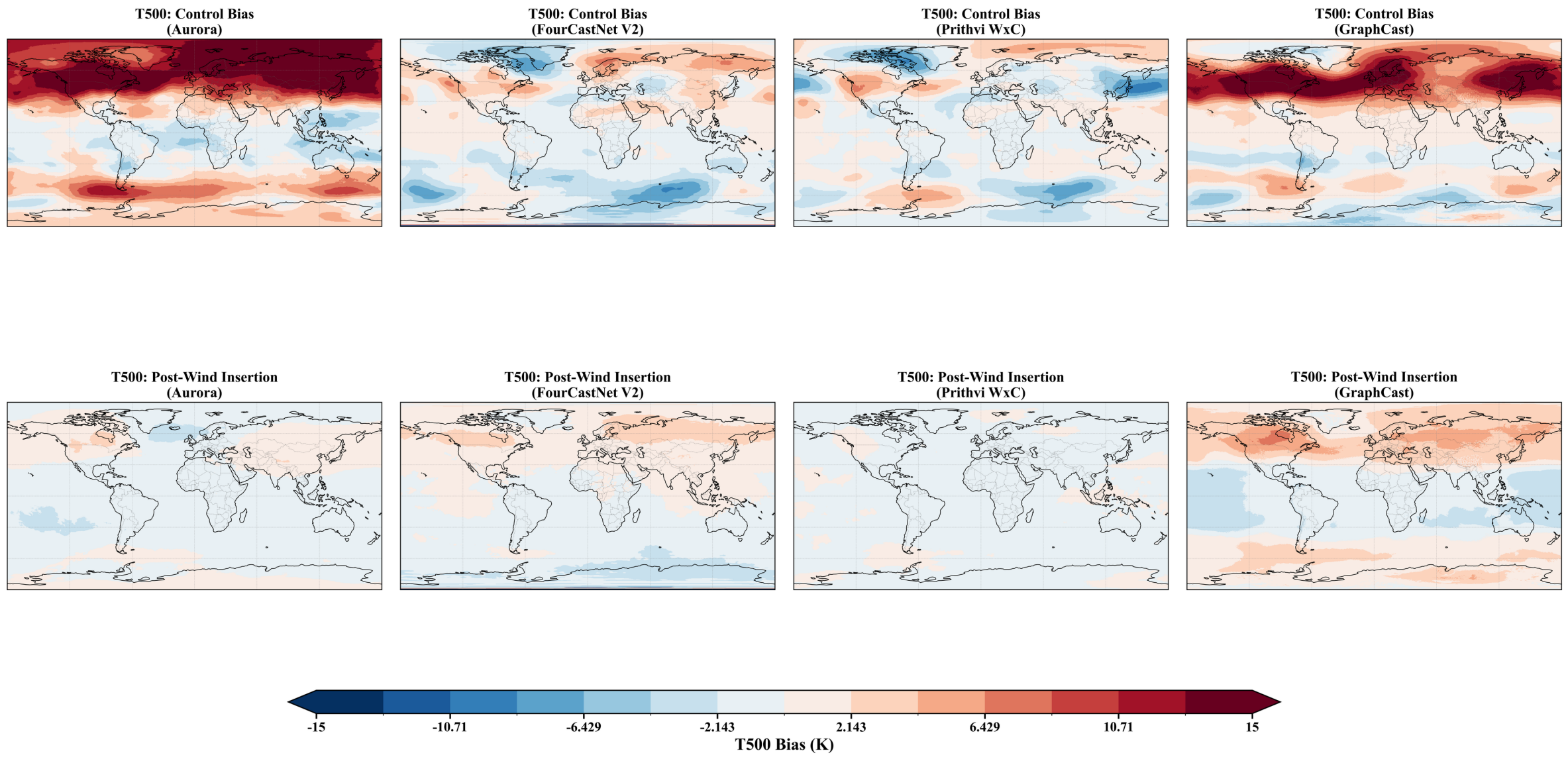

**Fig. 11.** Spatial bias in T500 (K) during the assimilation phase (days 30-60): each model's own free-running control bias (top row) versus its own post-wind-insertion bias (bottom row), for Aurora, FourCastNet V2, Prithvi WxC, and GraphCast (left to right).

#### 4.3.3 Impact of Wind Insertion on Geopotential

Wind and geopotential are dynamically linked through the horizontal momentum balance. In the free atmosphere, a corrected wind field should correspond to a corrected geopotential gradient through geostrophic balance (Holton & Hakim, 2013). Wind corrections also modify convergence, divergence, and vertical motion, redistributing atmospheric mass and allowing the geopotential field to adjust. A successful wind-to-geopotential response therefore requires the model to translate the imposed circulation into a dynamically consistent mass field. This relationship is generally strongest above the boundary layer, while friction, terrain, and surface processes make the response near 1000 hPa more complex.

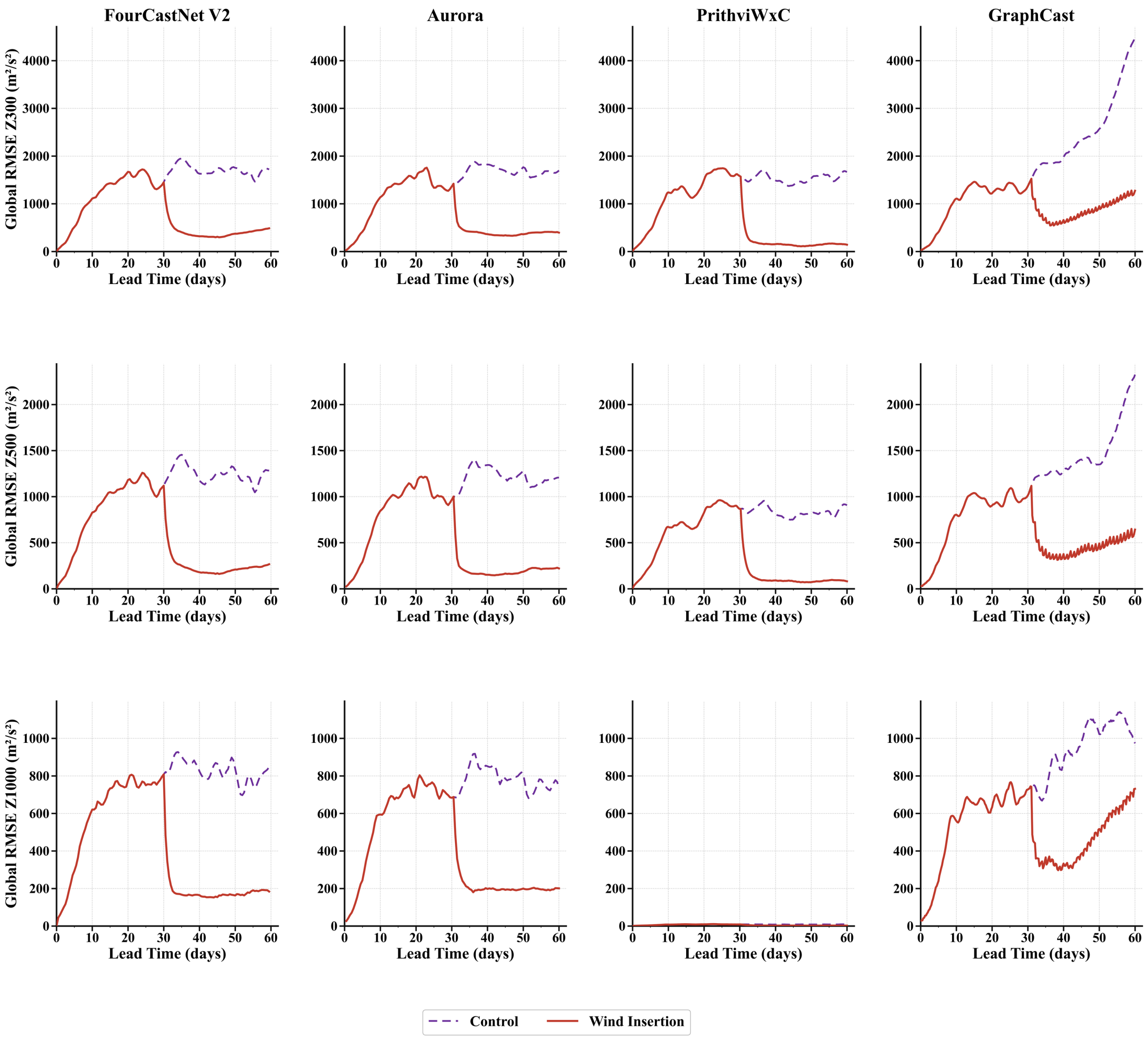

**Fig. 12.** Global RMSE of Geopotential (Z) at 300, 500, and 1000 hPa following wind insertion.

Before day 30, the control and insertion runs are identical. Once wind insertion begins, FourCastNet V2 and Aurora show immediate and sustained geopotential-error reductions at all three levels, indicating strong and persistent wind-mass coupling. Aurora produces a particularly smooth and stable adjustment, consistent with its multivariate three-dimensional latent representation (Bodnar et al., 2025), while FourCastNet V2's global spectral mixing (Pathak et al., 2022; Bonev et al., 2023; Li et al., 2021; Guibas et al., 2022) efficiently converts large-scale wind corrections into coherent geopotential-gradient adjustments. Prithvi WxC also shows strong physical consistency on the wind-to-geopotential pathway: its mature-phase, all-level wind-to-geopotential coupling score is 81.1% (Appendix B3, Fig. B4), and its latitude-resolved Z500 response after wind insertion (39.1% at 2N, rising to 75.5%, 87.3%, and 84.2% at 14N, 40N, and 60N) shows the same broad, latitude-tapering recovery seen in Aurora and FourCastNet V2. For Prithvi WxC specifically, this pathway can be traced to a concrete mechanism: wind insertion reduces its mature-phase surface-pressure and sea-level-pressure RMSE by 87.2% and 86.5% respectively (days 30-60, relative to control), more than double the 39.2% reduction achieved by temperature insertion alone (see “Impact of Temperature Insertion on Geopotential” above). Because geopotential height at any level is hydrostatically anchored to the surface-pressure/mass field rather than to the temperature profile alone, a wind-driven correction that reaches the surface-pressure field this much more effectively than a temperature-driven one is a plausible explanation for why wind insertion propagates into a larger geopotential-height correction for this model, even when the inserted temperature field is itself exact. GraphCast shows a strong initial adjustment but does not sustain it broadly across latitude: its Z500 response after wind insertion is strongly adverse at the two lowest latitude bands (−72.5% at 2N, −14.6% at 14N) before turning strongly positive at 40N and 60N (+76.9%, +75.9%), the same qualitative low-latitude weakness documented for GraphCast throughout this study. Prithvi WxC’s near-surface sigma-level geopotential-height curves (approximately 985 hPa, the native level closest to 1000 hPa) are a specific exception worth flagging rather than reading at face value: both its control and wind-insertion curves sit far below the other three models’ curves, even after the standard H-to-Z unit conversion (×g). Unlike FourCastNet V2, Aurora, and GraphCast, which all show clearly separated, physically substantial control-versus-insertion curves at 1000 hPa, Prithvi WxC’s curves at its approximately 985-hPa sigma level are nearly flat. This behavior should not be read as evidence of strong or weak wind–geopotential coupling for Prithvi WxC; it instead flags this diagnostic as needing a closer check of the underlying near-surface geopotential-height data before drawing any conclusion from it.

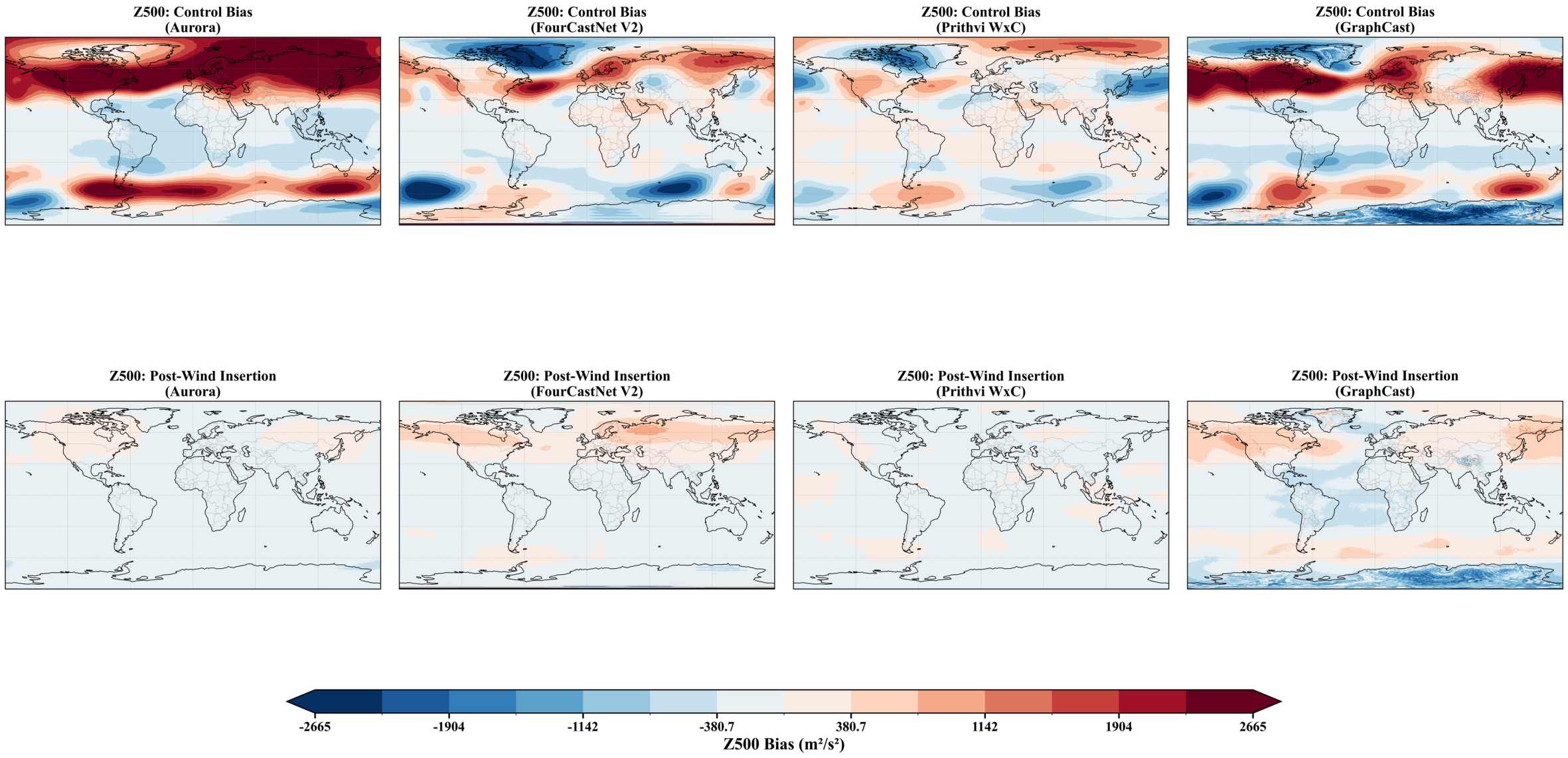

**Fig. 13.** Spatial bias in Z500 during the assimilation phase (days 30-60): each model's own free-running control bias (top row) versus its own post-wind-insertion bias (bottom row), for Aurora, FourCastNet V2, Prithvi WxC, and GraphCast (left to right).

Fig. 13 shows that wind insertion visibly flattens the Z500 bias field for all four models, including GraphCast, whose post-insertion bias is markedly reduced relative to its control-run pattern despite its weaker performance on other pathways.

4.3.4 Latitudinal response to global wind insertion

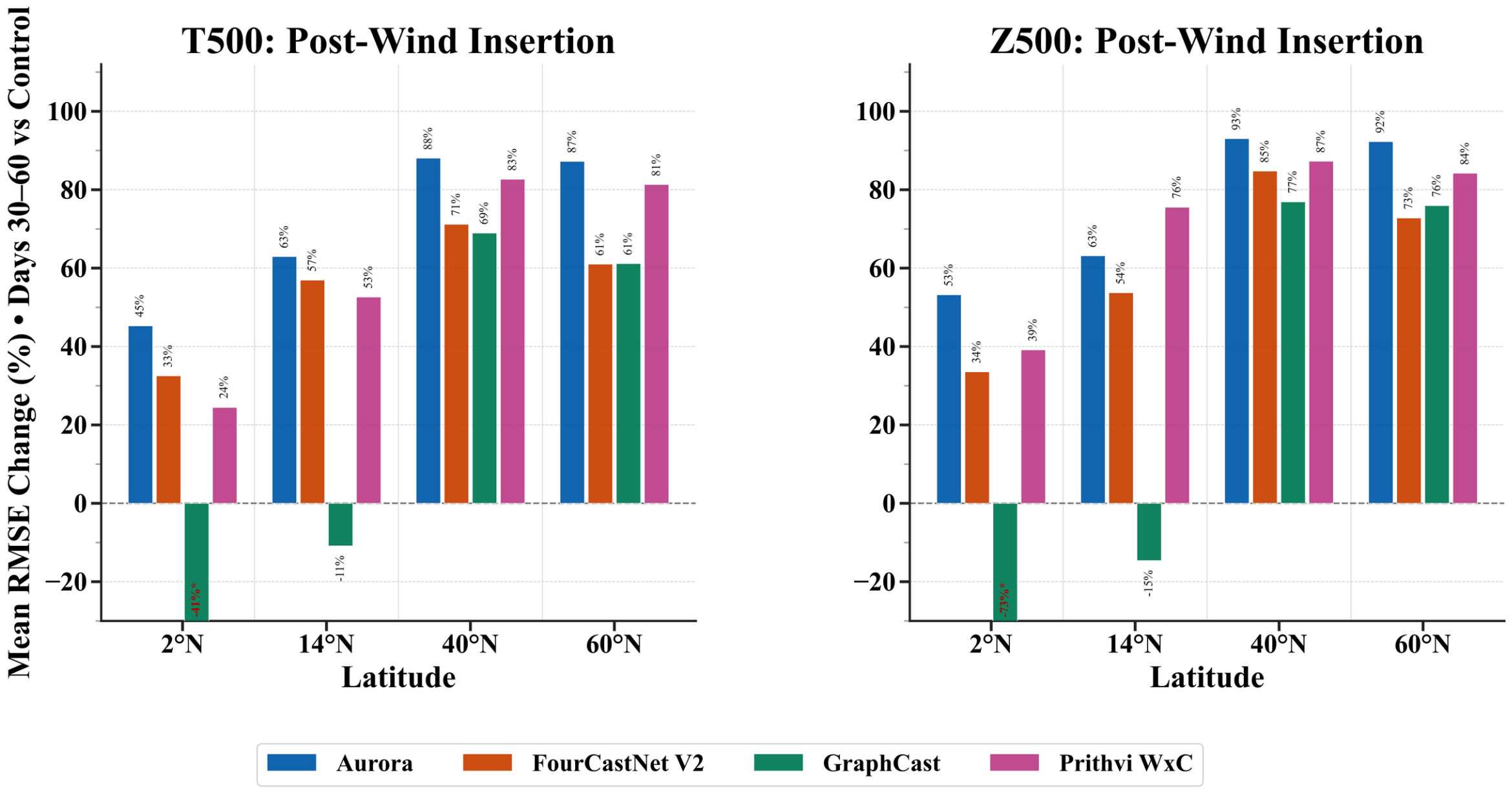

**Fig. 14.** Cross-variable RMSE response to Wind insertion during the assimilation phase (lead days 30-60). Each panel shows the mean percentage reduction in RMSE relative to the free-running control, averaged over days 30-60.

In the real atmosphere, wind corrections influence temperature through horizontal heat transport and circulation-driven vertical motion, while geopotential adjusts through dynamical mass–wind balance (Holton & Hakim, 2013). These relationships are generally stronger in the extratropics, where geostrophic and thermal-wind balance dominate, and more complex in the tropics, where divergent circulation and convection become important.

- Aurora: Aurora most consistently reproduces this expected latitude dependence. Its response strengthens outside the tropics, indicating that its multivariate three-dimensional representation (Bodnar et al., 2025) effectively propagates corrected circulation through horizontal and vertical atmospheric couplings.
- FourCastNet V2: FourCastNet V2 also shows strong extratropical adjustment. Its global spectral operator (Pathak et al., 2022; Bonev et al., 2023; Li et al., 2021; Guibas et al., 2022) appears to translate wind corrections into large-scale temperature and geopotential responses by representing synoptic circulation and thermal advection across broad spatial scales.
- GraphCast: GraphCast captures balanced wind–temperature coupling in the extratropics, but its response becomes less dynamically consistent near the equator, where weak Coriolis forcing limits geostrophic adjustment. Tropical wind insertion increases both temperature and geopotential error (Section 4.3.2-4.3.3), confirming that its learned dynamics do not adequately handle divergent circulation and other ageostrophic tropical processes.
- Prithvi WxC: Prithvi WxC shows a clear, latitude-tapering temperature and geopotential adjustment following wind insertion - weakest at 2N and strongest at 40-60N, the same qualitative pattern as Aurora and FourCastNet V2. Prithvi WxC's representation converts wind corrections into balanced mass-field and thermal adjustments through the horizontal pressure-gradient and Coriolis relationships required for geostrophic and thermal-wind balance; its mature-phase coupling scores (71.3% wind-to-temperature, 81.1% wind-to-geopotential) are the highest and second-highest, respectively, of all four models on these two pathways (Appendix B3, Fig. B4).

## 5. Correlated Observation-Error Perturbation Insertion

All insertion experiments described in the main text overwrite the model state with the error-free verifying analysis every six hours at every pressure level, an idealized, upper-bound configuration. Real observing systems never supply perfect data, so here we replace that idealized baseline with a Correlated OSSE Perturbation (COP) – a single, physically structured random field with specified horizontal and vertical correlation and, for wind, non-divergent structure; the full generation procedure is described in Section 3. Three models are evaluated under this scheme: FourCastNet V2, Aurora, and Prithvi WxC, retaining the same 60-day integration and 30-day spin-up protocols used elsewhere in this study; GraphCast is not included in this section, consistent with this experiment's three-model design.

### 5.1 Correlated Temperature Perturbation Insertion

Following the COP procedure described in Section 3, Fig. 15 shows the global RMSE time series of zonal wind under correlated-perturbation insertion, relative to each model's own free-running control, at the same three representative levels (300, 500, 1000 hPa). Recovery under correlated perturbation is compared numerically with the error-free insertion benchmark in the text.

#### 5.1.1 Effect on Wind

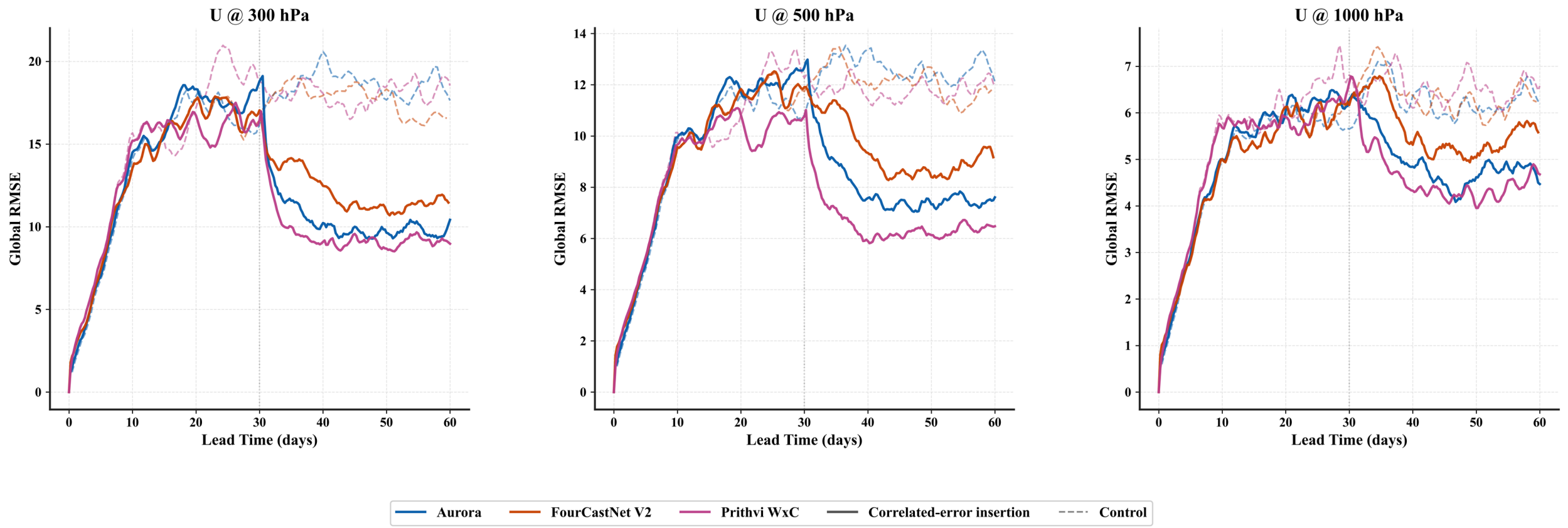

**Fig. 15.** Global RMSE of zonal wind (u) at 300, 500, and 1000 hPa following perturbed temperature insertion across three AI weather models (Aurora: blue; FourCastNet V2: orange; Prithvi WxC: magenta; solid: correlated-perturbation insertion; dashed: each model's own free-running control; GraphCast is not included in this section; see Section 5).

Fig. 15 (Global RMSE at U300/U500/U1000, all three models, correlated-perturbation insertion versus each model's own control) shows that once the perturbed-insertion curve breaks away from control around day 30, it settles into a clearly separated lower band for the rest of the window at every level, for all three models. Mature-phase (day 30-60) recovery under the correlated perturbation is essentially the same as under error-free insertion: FourCastNet V2 recovers 30.9% (U300), 21.9% (U500), and 11.6% (U1000), against an error-free benchmark of 29.0%, 18.6%, and 9.7%; Aurora recovers 43.2%, 36.0%, and 20.8%, against 43.9%, 37.4%, and 22.2%; and Prithvi WxC recovers 46.6%, 43.8%, and 29.1%, against 46.9%, 44.0%, and 29.7%. Every correlated-insertion value lies within 3.5 percentage points of its error-free counterpart, so the wind response is essentially insensitive to realistic, correlated observation error. Aurora's U300 panel additionally shows a distinct brief spike right at the injection transition before settling into the post-injection band - a transient specific to Aurora's onset behavior.

All three models show the same vertical structure already documented for the temperature-to-geopotential pathway: recovery is largest in the upper troposphere (300 hPa) and falls off toward 1000 hPa, consistent with the geostrophic/thermal-wind relationship being most directly expressed near jet level and increasingly complicated by boundary-layer and surface effects closer to the ground. This holds for Prithvi WxC as well (46.6% at 300 hPa falling to 29.1% at 1000 hPa). Of the two ERA5-trained models, Aurora shows the stronger wind response at every level tested under this correlated design (U300 43.2% versus 30.9% for FourCastNet V2; V300 39.6% versus 31.1%). Prithvi WxC's U300 recovery (46.6%) is slightly ahead of Aurora's, and its V300 recovery (39.7%) essentially matches Aurora's and exceeds FourCastNet V2's: on the numbers available here, Prithvi WxC's wind response is the strongest of the three at every level tested (for example, U500 43.8% versus 36.0% for Aurora and 21.9% for FourCastNet V2), though only marginally at V300.

### 5.1.2 Effect on Geopotential (Z)

Fig. 16 (Global RMSE at Z300/Z500/Z1000, all three models, correlated-perturbation insertion versus each model's own control) shows FourCastNet V2 and Aurora both transitioning from control to a lower band over roughly the first ten days after injection begins, while Prithvi WxC drops almost immediately; all three remain in this lower band for the rest of the window at every level - the same pattern documented for wind. Aurora's geopotential recovery (Z300 49.7%, Z500 41.0%) exceeds FourCastNet V2's (45.2%, 33.4%). Prithvi WxC's geopotential recovery under this experiment is 57.7% (Z300), 60.1% (Z500), and 60.7% (Z1000), the highest of the three models at every level. Each model's COP geopotential response tracks its own error-free benchmark closely at every level sampled: within 2.3 percentage points for Aurora, 3.4 for FourCastNet V2, and 2.2 for Prithvi WxC (see Section 5.1.3 for the latitude-resolved comparison at Z500).

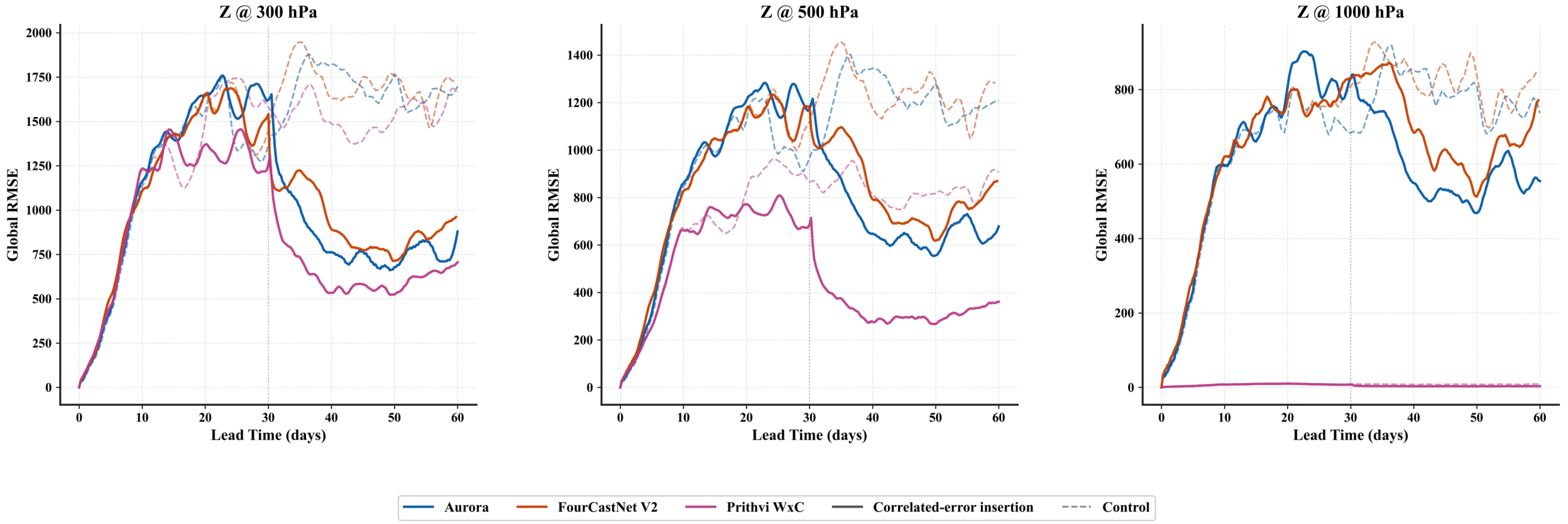

**Fig. 16.** Global RMSE of geopotential (z) at 300, 500, and 1000 hPa following perturbed temperature insertion across three AI weather models (solid: correlated-perturbation insertion; dashed: each model's own free-running control). Prithvi WxC geopotential height (m) is multiplied by $g = 9.80665$ m $s^{-2}$ to match the other models' geopotential ($m^2 s^{-2}$); its 1000 hPa curve lies near zero because Prithvi WxC's lowest level is terrain-following. GraphCast is not included in this section; see Section 5.

### 5.1.3 Latitudinal Response to Correlated Temperature-Perturbation Insertion

**Fig. 17.** Mean RMSE change (%) relative to control over days 30-60 for Z500, U500, and V500 at four latitude slices (2N, 14N, 40N, 60N), correlated temperature-perturbation insertion, FourCastNet V2 and Prithvi WxC. Aurora is omitted because a matching latitude-resolved control was not available for its COP runs.

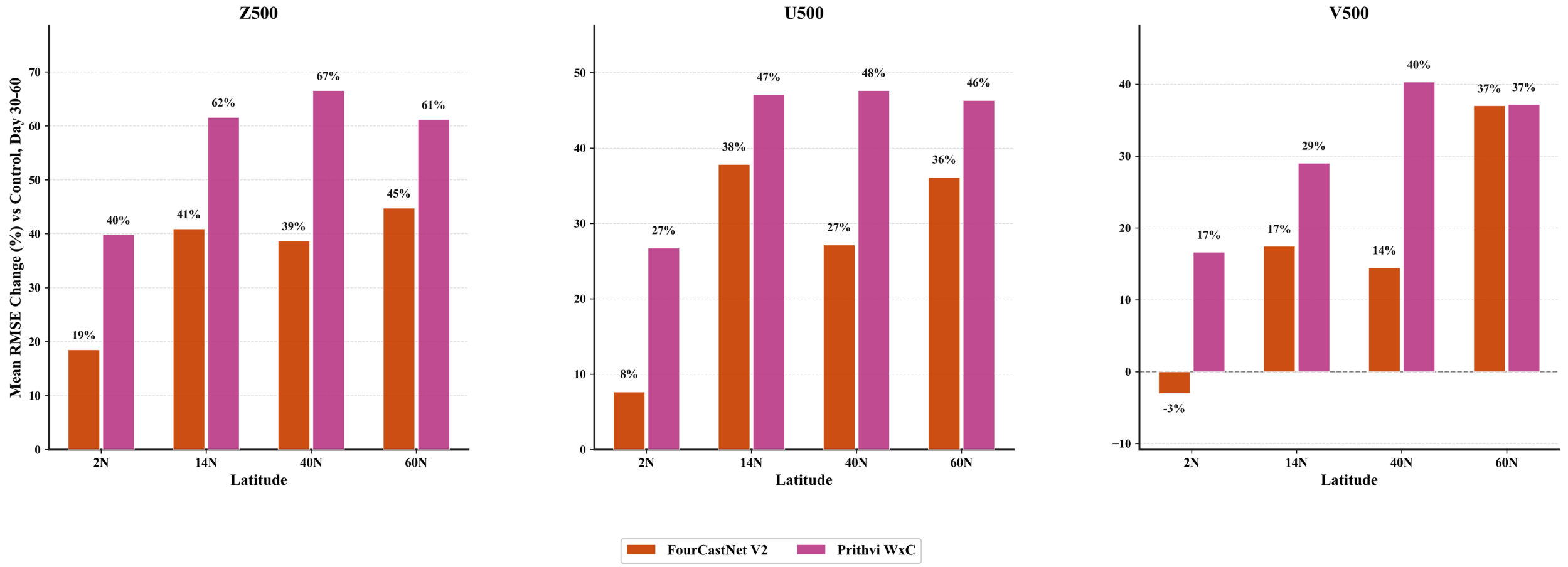

Figure 17 reports mean RMSE change (%) relative to control over days 30-60, at four latitude slices (2N, 14N, 40N, 60N), for the cross-variable wind and geopotential response to correlated temperature-perturbation insertion in FourCastNet V2 and Prithvi WxC; the error-free values quoted for comparison are computed the same way from the error-free temperature-insertion runs.

For Z500, correlated-insertion recovery stays close to error-free recovery at most latitudes. For FourCastNet V2 the gap is at most 3.0 percentage points from 2N to 40N (18.5%, 40.9%, and 38.7% versus 18.6%, 37.9%, and 39.8% error-free), while at 60N the correlated run recovers 44.8% against 36.8% error-free. Prithvi WxC's Z500 bars show close pairing at every latitude: correlated-versus-error-free gaps of 1.6 (2N), 3.8 (14N), 1.0 (40N), and 1.7 (60N) percentage points, with the largest gap at 14N. The magnitude of Prithvi WxC's correlated-insertion recovery is also large and latitude-consistent (39.8% at 2N rising to 66.6% at 40N), exceeding FourCastNet V2's at every latitude.

For U500, Prithvi WxC's correlated-insertion recovery (26.7% to 47.6% across the four bands) is close to its own error-free counterpart (30.8% to 52.2%) at every latitude, with gaps of 2.7-5.1 percentage points. FourCastNet V2's U500 recovery is weakest at 2N (7.7%) and reaches 27.1-37.8% poleward; it is within 4.2 percentage points of its error-free counterpart from 2N to 40N, but about 10 points higher at 60N (36.1% versus 26.2%). Both models therefore show the weak-tropics, stronger-extratropics pattern expected from geostrophic adjustment, with a response that is largely insensitive to the correlated noise. V500's response is smaller in magnitude than U500's or Z500's for both models at nearly every latitude, and FourCastNet V2's V500 response is slightly negative at 2N (-3.0%).

### 5.2 Correlated Zonal and Meridional Perturbation Insertion

Following the same COP procedure (Section 3), but applied to zonal and meridional wind instead of temperature.

#### 5.2.1 Effect on Temperature (T)

Fig. 18 (Global RMSE at T300/T500/T1000, all three models, correlated-perturbation insertion versus each model's own control) shows FourCastNet V2 and Aurora both dropping sharply away from control at day 30 and remaining in a lower band for essentially the whole remaining window at every level. Aurora's panels also show a sharp spike right at the very end of the window (day ~59-60), which is absent from its error-free wind-insertion run; because it is confined to the last day, it has little effect on the day 30-60 averages. Both models peak at 500 hPa (Aurora 65.1%, FourCastNet V2 67.8%) rather than at jet level, but differ sharply near the surface: FourCastNet V2 retains 47.1% at 1000 hPa (62.4% at 300 hPa), while Aurora falls to 14.9% (47.1% at 300 hPa). Prithvi WxC's T300/T500/T1000 panels show the same sharp drop away from control: mature-phase recovery is 70.3% (T300), 77.1% (T500), and 65.0% (T1000), essentially matching its own error-free wind-to-temperature response at the same levels (71.5%, 78.0%, 65.7%) to within about one percentage point, and exceeding both FourCastNet V2 and Aurora at every level. Aurora and FourCastNet V2 likewise stay within 2.6 percentage points of their error-free benchmarks at every level. Wind insertion under this correlated design provides Prithvi WxC's temperature field a substantial benefit.

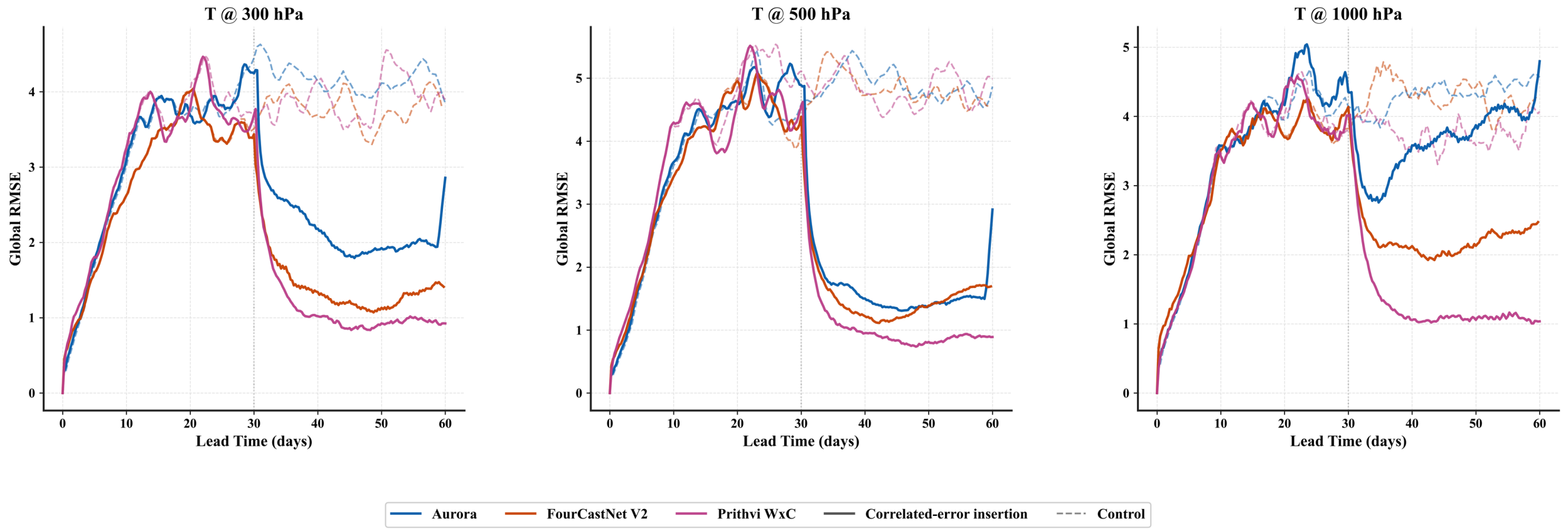

**Fig. 18.** Global RMSE at T300/T500/T1000 post perturbed wind insertion (solid: correlated-perturbation insertion; dashed: each model's own free-running control).

#### 5.2.2 Effect on Geopotential (Z)

Fig. 19 (Global RMSE at Z300/Z500/Z1000, all three models, correlated-perturbation insertion versus each model's own control) shows the same pattern as before: FourCastNet V2 and Aurora both drop sharply at day 30 and remain in a lower band for nearly the entire window at every level, with Aurora again showing the end-of-window spike noted above. FourCastNet V2's and Aurora's wind-to-geopotential response (71.5-81.1% across levels) is stronger than either model's own temperature-to-geopotential response - momentum balance reaching geopotential

more directly than the hydrostatic, temperature-mediated pathway, for both models, under noisy as well as exact insertion. Prithvi WxC's wind-to-geopotential response under this experiment is 85.8% (Z300), 84.6% (Z500), and 64.7% (Z1000) - above the 71.5-81.1% range of FourCastNet V2 and Aurora at Z300 and Z500, but below it at Z1000. Prithvi WxC's own wind-to-geopotential response is also stronger than its temperature-to-geopotential response at every level under the correlated-perturbation design (85.8% vs. 57.7% at 300 hPa; 84.6% vs. 60.1% at 500 hPa; 64.7% vs. 60.7% at 1000 hPa) - a real asymmetry between the two pathways.

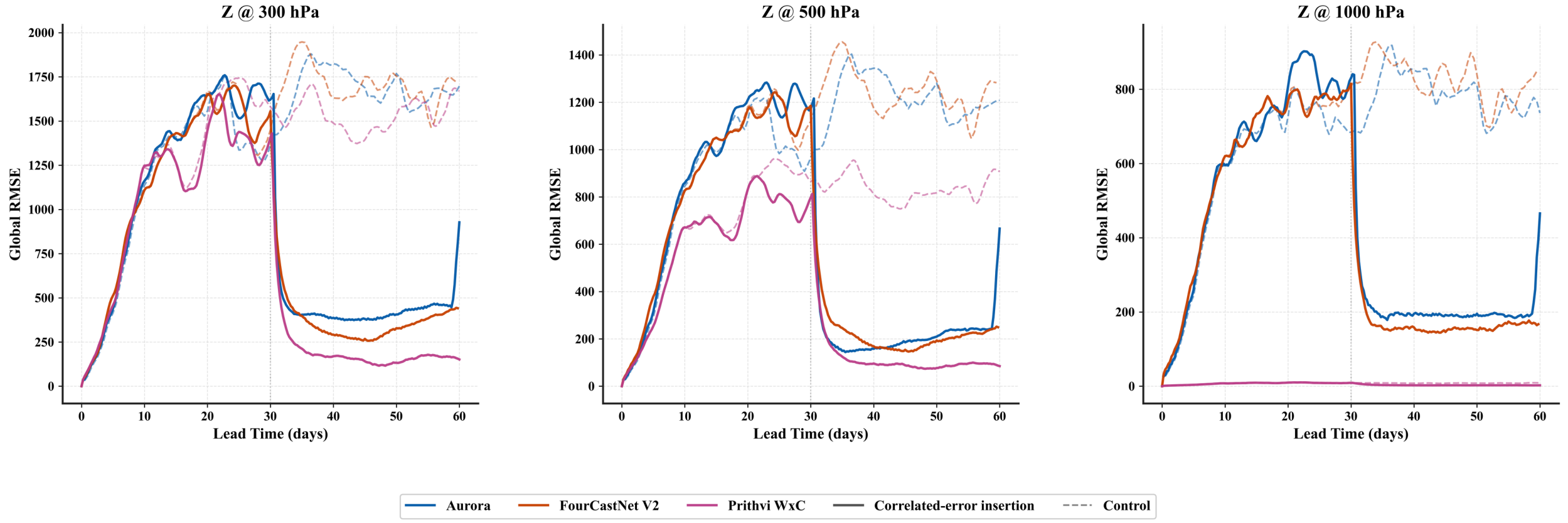

**Fig. 19.** Global RMSE at Z300/Z500/Z1000 post perturbed wind insertion (solid: correlated-perturbation insertion; dashed: each model's own free-running control; Prithvi WxC geopotential converted to m² s⁻² as in Fig. 16).

## 6. Discussion

### a. Limitations

All insertion experiments in this study use a single deterministic 60-day autoregressive rollout per model and configuration, initialized from one reference date. Consequently, the reported RMSE improvements and cross-variable coupling percentages are point estimates from a single realization rather than statistics averaged across multiple initial conditions, seasons, or ensemble members. Day-to-day variability within the mature phase is present but is not formally quantified. Small differences between models, on the order of a few percentage points, should therefore be interpreted cautiously, as they may partly reflect the specific rollout period rather than a robust architectural distinction. Larger differences that recur across multiple variables, latitude bands, and diagnostic classes provide stronger evidence of systematic behavior within the evaluated period, although these diagnostics are not independent realizations. GraphCast's weak and frequently negative low-latitude coupling scores, together with weak or negative scores across broader regions for several pathways, illustrate such a recurring pattern in this study. Future work should repeat the core insertion experiments across multiple initial conditions, seasons, and weather regimes to quantify run-to-run variability and strengthen conclusions about model-to-model differences.

The present experiments provide a controlled first step toward understanding how AI weather models respond to observational constraints. However, the use of complete, error-free atmospheric fields differs considerably from operational observing conditions. A natural extension would introduce satellite-derived observations with realistic spatial coverage, uncertainty, and missing data. Combining these observations with model forecasts would allow the resulting systems to be evaluated for their ability to integrate observational information rather than solely for their forecasting performance.

An additional limitation concerns the vertical extent of temperature insertion in GraphCast. Inserting temperature at all 37 native pressure levels caused the rollout to become unstable. The reported temperature-insertion results were obtained only after excluding the 11 upper-atmospheric levels from 1–100 hPa and restricting insertion to the remaining 26 levels from 125–1000 hPa. These results therefore describe GraphCast's response to a restricted temperature-insertion configuration, rather than insertion across its full vertical domain. The instability under full-level insertion is itself a limitation of the model's response to the imposed constraints in this experimental setup, and its underlying cause requires further investigation.

Finally, direct field replacement does not account for observational and forecast uncertainty in the way that formal data assimilation does. Future work could replace this approach with established methods, such as the ensemble Kalman filter (EnKF), local ensemble transform Kalman filter (LETKF), or four-dimensional variational assimilation (4D-Var), adapted for use with AI weather models. These methods would combine observations with model forecasts while accounting for their respective uncertainties, allowing a more realistic assessment of the models' response to sparse and noisy observations.

**b. Conclusion**

This study introduced a direct-insertion ML-OSSE framework and applied it to four architecturally distinct AI weather models: the spectral FourCastNet V2, the graph-based GraphCast, and the foundation-model transformers Aurora and Prithvi WxC. The goal was to test whether models trained only to minimize forecast error have also learned the physical balance relationships (hydrostatic, geostrophic, thermal-wind, moisture coupling) that a data-assimilation system depends on, following and extending Charney et al.'s (1969) temperature-insertion design to wind insertion as well. Temperature insertion recovers the hydrostatic (T→Z) pathway more reliably than any other pathway tested: Aurora and FourCastNet V2 cut Z500 RMSE by 66-82% and 19-40% across latitude, and Prithvi WxC by 41-68% (Fig. 7); GraphCast is the clear outlier, adverse at three of the four latitude bands sampled (-53% to -162%). At the equator specifically, Aurora and Prithvi WxC, the two foundation-model architectures, retain a clearly positive temperature-to-wind response (≈52% and ≈31% RMSE improvement in U500 at 2°N; Fig. 6), far above FourCastNet V2 (≈6%) and GraphCast (actively adverse, -8.5%). This is a genuine departure from the common assumption, traced to Halem et al. (2000), that this coupling nearly vanishes near the equator; it instead confirms Charney et al.'s (1969) original, all-latitude conjecture and Ghil et al.'s (1977) theoretical result, in a new class of learned nonlinear dynamical systems. We also find that a simple, intuitive asymmetry (constraining temperature/geopotential controls wind more effectively than the reverse) is not architecture-independent: Aurora's wind-to-temperature (68.9%) and wind-to-geopotential (79.0%) coupling scores both exceed its own temperature-to-wind scores (38.6% U, 21.4% V; Appendix B3, Fig. B4), the opposite direction.

Wind insertion is the strongest overall pathway in the study. Averaged across the full cross-variable coupling matrix (Appendix B3, Fig. B4), Prithvi WxC leads (52.2%), followed by Aurora (46.2%), FourCastNet V2 (42.6%), and GraphCast (20.5%). For Prithvi WxC this traces to a concrete mechanism: wind insertion cuts its surface-pressure and sea-level-pressure RMSE by 87.2%/86.5%, more than double temperature insertion's 39.2%/39.3%, reaching the mass-field boundary condition that anchors hydrostatic geopotential height more effectively than a temperature correction alone. Aurora shows the broadest, most latitude- and terrain-consistent recovery of any single model and remains the strongest current OSSE-screening candidate once consistency, not just peak coupling strength, is weighed in (Appendix B2/B3). GraphCast again lags, frequently negative in the deep tropics; its own control-run error is already growing rapidly by day 40 (Section 4.1), which complicates a clean read of its mature-phase results specifically.

This is a deliberately idealized, best-case test: error-free truth overwritten at every level, with no observation operator, sparse sampling, or background-error weighting as in real 3D/4D-Var (Kalnay, 2003). It establishes an upper bound on how much physical information each architecture can extract from perfect observations, not a forecast of real observing-system impact. The natural next step is coupling these models to genuine assimilation machinery, sparse pseudo-observations, and explicit observation operators (Lieberman et al., 2026) to test whether this physical consistency survives realistic sampling and noise (Tian et al., 2024; Xu et al., 2025; Fan et al., 2025; Arcucci et al., 2026).

**Inclusion in Global Research.**

This study uses publicly available reanalysis products (ERA5, MERRA-2, CAMS) and openly released, pretrained AI weather models; it did not involve fieldwork, human subjects, or data collection in partnership with local communities, so no additional permits, authorizations, or community-engagement considerations apply.

**Materials and Methods.**

During manuscript preparation, ChatGPT (OpenAI) was used to assist with writing, organizing, and refining the text for clarity and readability. Claude Code (Anthropic) was used to assist with formatting and preparing the manuscript in Microsoft Word. The authors reviewed and edited the resulting text and formatting and took full responsibility for the accuracy and integrity of the final manuscript.

**Open Research**

1. **Software Availability**

The truth-insertion OSSE code for each model, including the full modified model source, experiment drivers and job scripts, preprocessing and analysis scripts, is archived on Zenodo: Aurora_OSSE v1.0.0 (Diggikar et al., 2026a; https://doi.org/10.5281/zenodo.24092590), FourCastNetv2_OSSE v1.0.0 (Diggikar et al., 2026b; https://doi.org/10.5281/zenodo.24092592), GraphCast_OSSE v1.0.0 (Diggikar et al., 2026c; https://doi.org/10.5281/zenodo.24092594), PrithviWxC_OSSE v1.0.0 (Diggikar et al., 2026d; https://doi.org/10.5281/zenodo.24092596), and OSSE_CrossModel_Analysis v1.0.0 (Diggikar et al., 2026e; https://doi.org/10.5281/zenodo.24092598).
They are developed openly at https://github.com/yashdiggikar under the MIT and Apache-2.0 licenses. The pretrained models used are publicly available from their developers: Aurora (Bodnar et al., 2025; https://huggingface.co/microsoft/aurora), FourCastNet v2-small (Bonev et al., 2023; via NVIDIA earth2mip), GraphCast (Lam et al., 2023;

https://github.com/google-deepmind/graphcast), and Prithvi WxC (Schmude et al., 2024; https://huggingface.co/ibm-nasa-geospatial/Prithvi-WxC-1.0-2300M).

## 2. Data Availability

The initial conditions and verification "truth" are the ERA5 reanalysis (Hersbach et al., 2020; Copernicus Climate Change Service Climate Data Store, https://doi.org/10.24381/cds.bd0915c6 for pressure levels and https://doi.org/10.24381/cds.adbb2d47 for single levels) and MERRA-2 (GMAO, 2015; NASA GES DISC).

The derived results that support the findings (RMSE time series, regime-stratified RMSE, spatial bias fields, and figure data) are archived on Zenodo (Diggikar et al., 2026f; https://doi.org/10.5281/zenodo.24092684). Raw model forecasts (about 15-80 GB per experiment) are available from the corresponding author on request.

**Acknowledgments.**

This work was supported by the NASA Earth Science Technology Office (grant 80NSSC22K1405). Experiments were conducted using the UMBC High-Performance Computing Facility (CHIP) and the infrastructure of the NSF Center for Accelerated Real-Time Analytics (CARTA) at UMBC. We thank Dr. Karuna Pande Joshi, UMBC Director of CARTA, for her support of the research infrastructure used in this work. We are grateful to Dr. Sujit Roy of NASA's Marshall Space Flight Center and Professor Mark A. Cane of Columbia University for reviewing the manuscript and for their thoughtful comments and suggestions, which helped strengthen the paper.

# APPENDIX

## Appendix A: Extended Insertion Experiments

### A.1 Regionalized Truth Insertion

This extension of the main OSSE design inserts truth only over land or ocean, using the ERA5 or MERRA-2 land–sea mask for the corresponding experiments.

- Models: Aurora, FourCastNet V2 (FCN), Prithvi WxC, and GraphCast for temperature insertion (Fig. A1, T500 panel). GraphCast is excluded from the Wind500 panel because its land/ocean wind-insertion data are unavailable.
- Experiments: Land-T, Ocean-T, Land-UV, and Ocean-UV.
- Protocol: 60-day rollout; free-running on days 1–30, with masked truth insertion on days 30–60.
- Metric: Cosine-latitude-weighted RMSE improvement relative to Control for T and U at 500 hPa, evaluated over the final 20-day mature phase (days 41–60).
- Question: Does correcting land or ocean improve forecasts in the uncorrected region, how far do improvements extend, and why does this range differ by model architecture?

Ocean insertion outperforms land insertion for temperature in all four models and for wind in all three with available data (Fig. A1). This may reflect the ocean's larger area and generally smoother, lower-variance atmospheric state, with fewer sharp orographic gradients, allowing corrections to propagate more easily. For Land-T, Prithvi WxC performs weakest: GraphCast achieves 44.0%, Aurora 42.3%, FCN 24.1%, and Prithvi WxC 23.4%. For Ocean-T, Prithvi WxC leads at 63.1%, narrowly ahead of Aurora and GraphCast (both 58.9%) and FCN (55.9%). FCN achieves the strongest ocean wind-insertion improvement (77.3%), exceeding the other two models with wind data and every other value in the chart. Overall, wind insertion yields weaker improvements than temperature insertion over both land and ocean, likely because the finer-scale spatial structures of zonal and meridional winds are harder to correct and propagate.

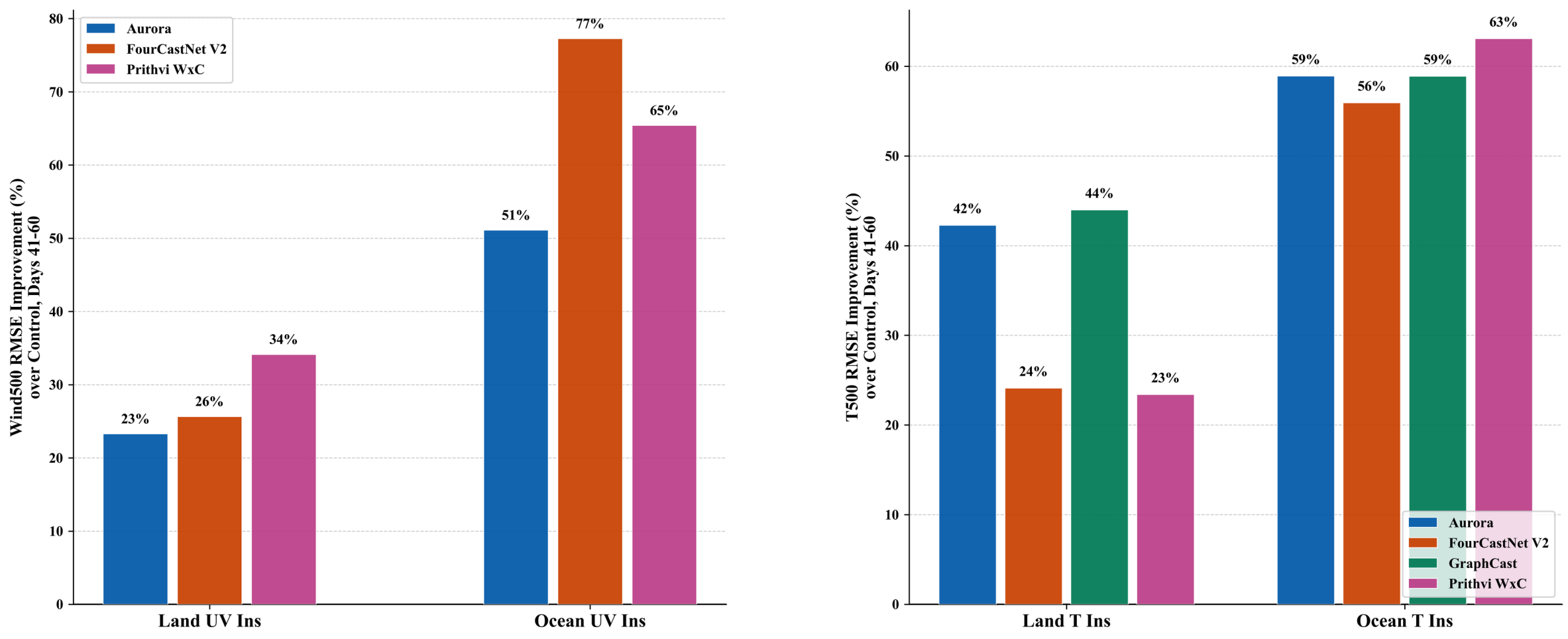

**Fig. A1.** T500, Wind500 RMSE improvement (%) over Control, days 41-60, for land- vs. ocean-masked temperature insertion for 4 models and wind insertion for 3 models (GraphCast excluded from the Wind500 panel).

- **All models:** A persistent dark band over the masked region across all three snapshots indicates sustained agreement with truth beyond the injection instant.
- **FCN:** Patchier, noisier errors outside the mask than Aurora or Prithvi, potentially consistent with diffuse error spreading through its spectral SFNO representation (Bonev et al., 2023).
- **Prithvi:** Widest dark halo beyond the mask, especially from Ocean-T into coastal land.
- **Aurora:** Large, sharp-edged synoptic error patches persist across all three days, potentially reflecting coherent mixing within Swin windows (Bodnar et al., 2025).

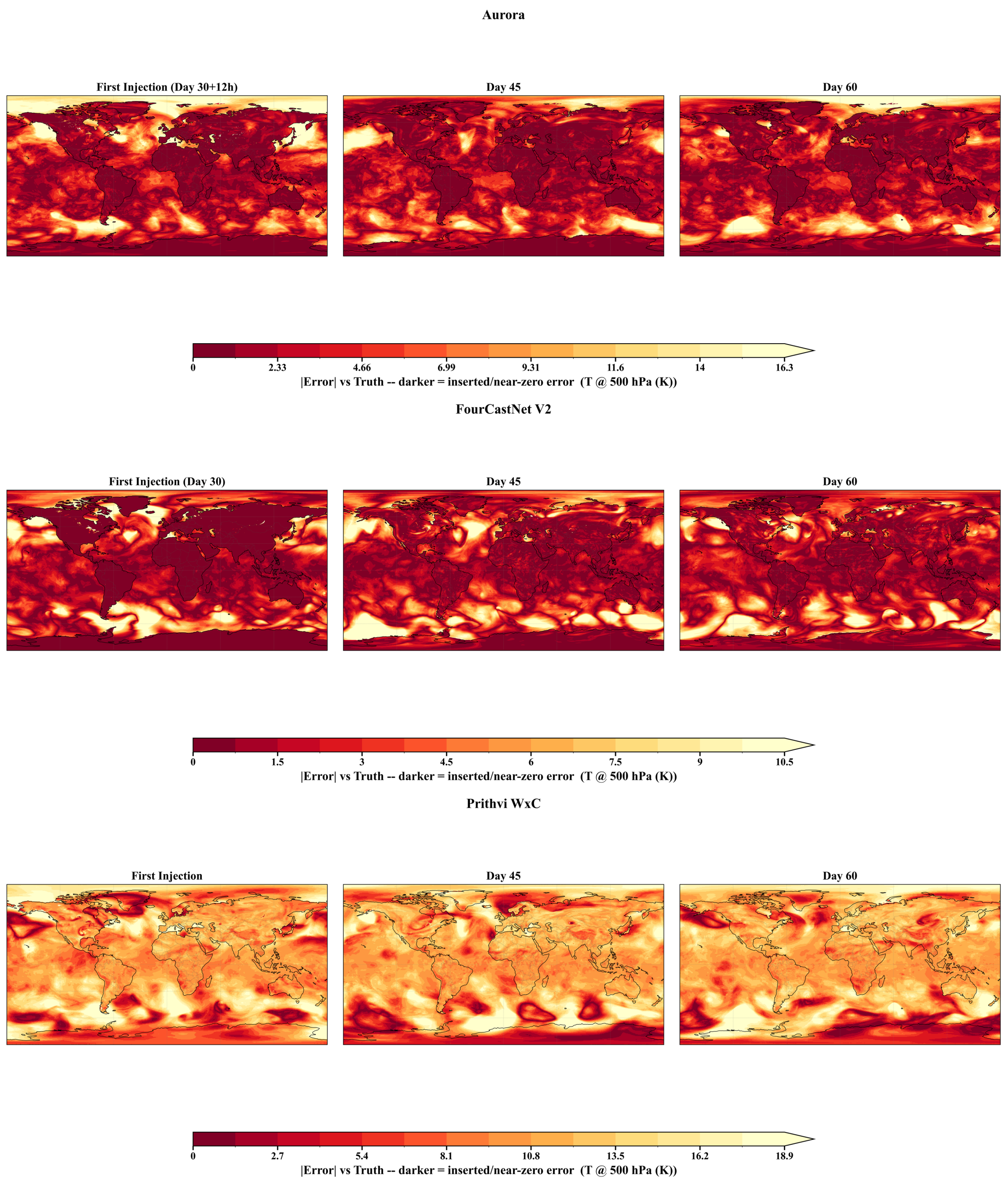
Aurora
First Injection (Day 30+12h)
Day 45
Day 60
0
2.33
4.66
6.99
9.31
11.6
14
16.3
|Error| vs Truth -- darker = inserted/near-zero error (T @ 500 hPa (K))
FourCastNet V2
First Injection (Day 30)
Day 45
Day 60
0
1.5
3
4.5
6
7.5
9
10.5
|Error| vs Truth -- darker = inserted/near-zero error (T @ 500 hPa (K))
Prithvi WxC
First Injection
Day 45
Day 60
0
2.7
5.4
8.1
10.8
13.5
16.2
18.9
|Error| vs Truth -- darker = inserted/near-zero error (T @ 500 hPa (K))

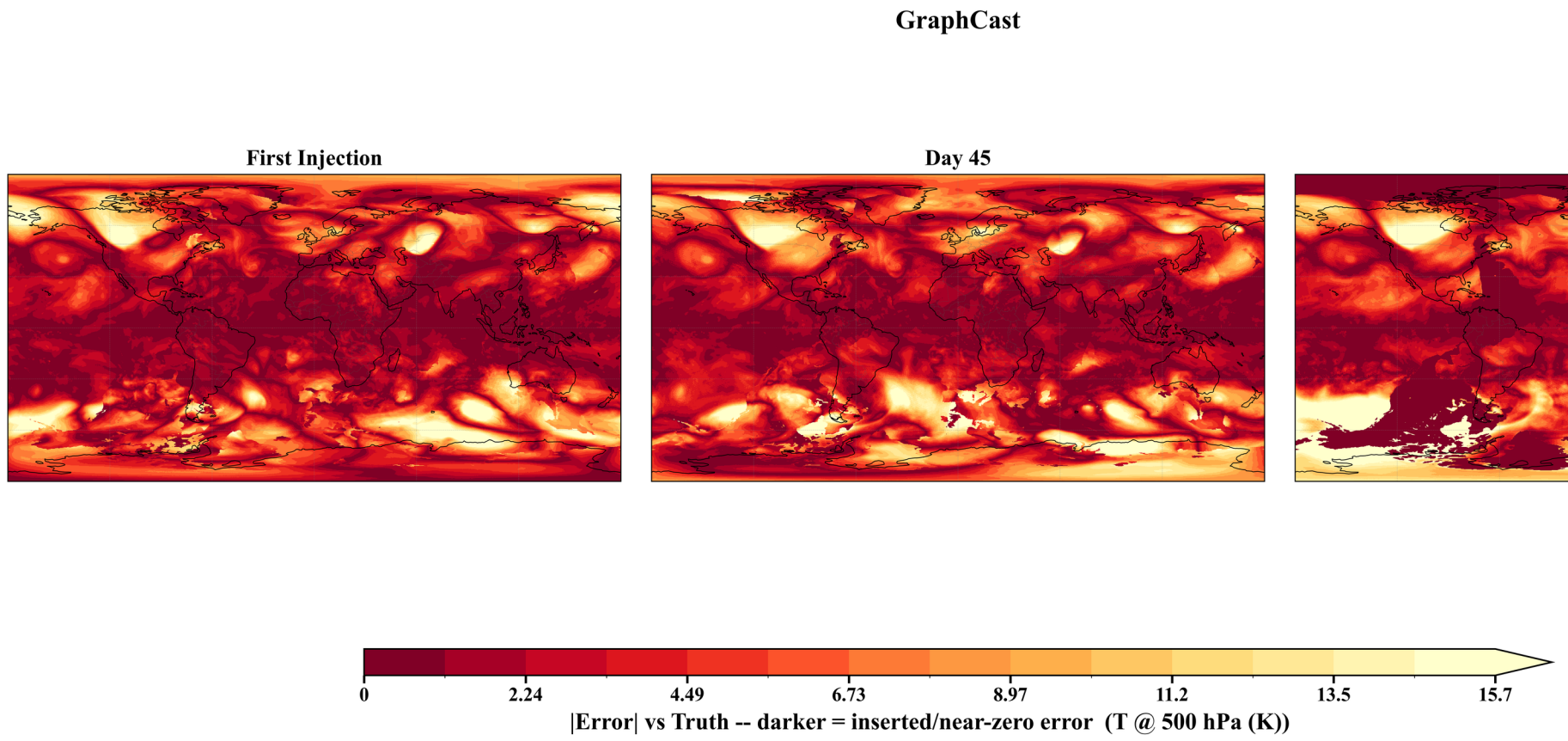

**Fig. A2.** |forecast − truth| error field for land-masked temperature insertion at Day 30, 45, and 60, with the darker masked-region band persisting through the full mature phase for all four models.

## Spillover-distance decay

This diagnostic measures how truth-insertion benefits extend beyond the mask edge with great-circle distance. Each model is evaluated on its native grid using its own land–sea mask; masks are not shared. A cKDTree on unit-sphere coordinates gives each cell's great-circle distance to the nearest inserted cell. Evaluation excludes the inserted region, whose near-zero errors would otherwise dominate. Non-inserted cells are grouped into 15 distance bins, from approximately 100 km near the boundary to several thousand kilometres.

Within each bin, cosine-latitude-weighted squared errors are pooled across space and time over days 41–60. RMSE is the square root of total weighted squared error divided by total weight, rather than an average of per-timestep RMSEs. Percentage improvement is 100 × (Control RMSE − experiment RMSE) / Control RMSE.

Matching land- and ocean-masked spillover runs are available for all four models, including GraphCast (Figs. A4–A5). Near-boundary (0–100 km) improvements range from 2.3% (Prithvi WxC, Land-T) to 62.0% (Aurora, Land-T). All models show positive corrections, indicating short-distance coupling between land and ocean, although Prithvi WxC responds much more weakly (2.3% Land-T; 5.8% Ocean-T) than the other three models (15.6–62.0%).

Land-to-ocean propagation is evaluated to 5000 km; ocean-to-land propagation has too few land cells beyond approximately 1700–1900 km. This reflects geography rather than an analysis limitation: no land location lies more than approximately 2000 km from an ocean cell, whereas ocean interiors can lie much farther from land.

A shift from positive to negative improvement indicates net harm, rather than merely no benefit, consistent with an insertion-induced, physically inconsistent discontinuity propagating outward. Prithvi WxC's Land-T curve crosses from positive to negative twice within the sampled range, rather than decaying only once.

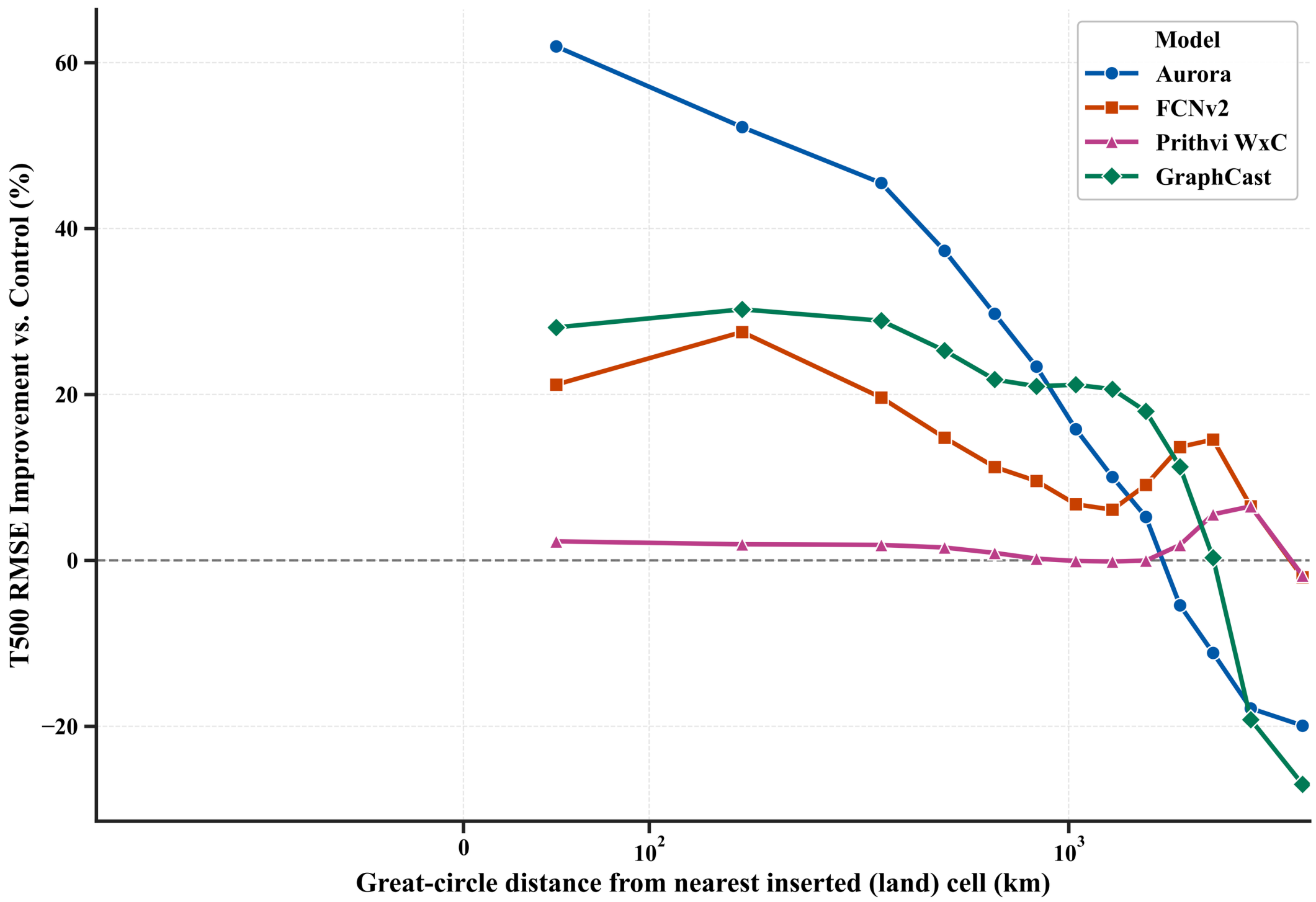

**Fig. A3.** T500 RMSE improvement (%) vs. Control as a function of great-circle distance from the nearest land-inserted cell, evaluated over ocean, days 41–60, for all four models.

T500 improvement falls from its near-boundary peak toward zero, and for three of the four models below zero, as distance from the nearest land-inserted cell increases.

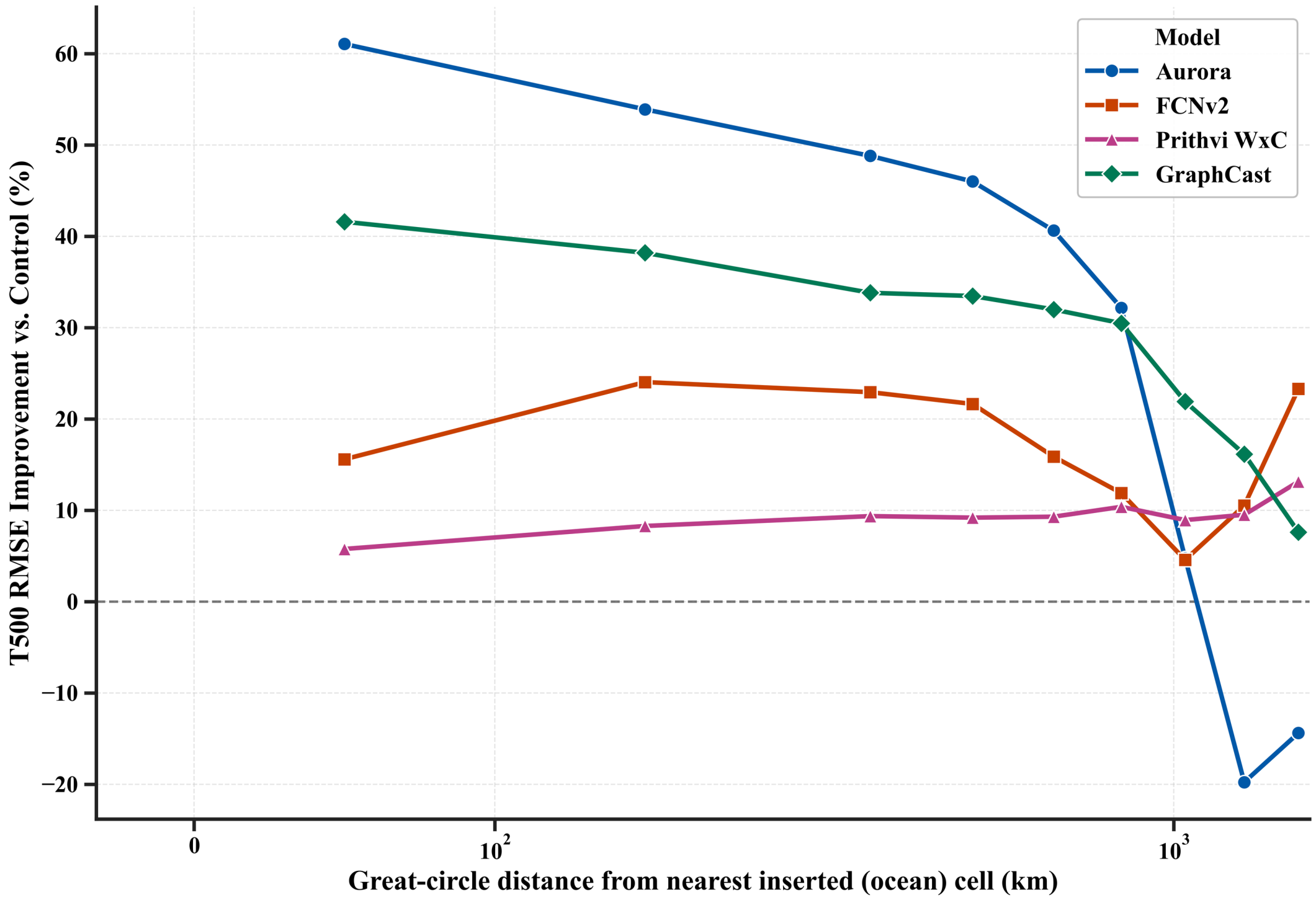

**Fig. A4.** T500 RMSE improvement (%) vs. Control as a function of great-circle distance from the nearest ocean-inserted cell, evaluated over land, days 41–60, for all four models.

Fig. A4 shows the equivalent decay for ocean-inserted cells evaluated over land, following the same distance-binned convention as Fig. A4.

## A.2 Temperature and Wind Combined Insertion

- Design: Extend the single-variable OSSE by simultaneously inserting truth for temperature and zonal/meridional winds.
- Models: FourCastNet V2, Aurora, and Prithvi WxC for both diagnostics; GraphCast for Z500 only because its combined-insertion run lacks humidity output.
- Objective: Test whether combined insertion outperforms the stronger single-variable experiment for unconstrained Z500 (hydrostatic/geostrophic response) and Q850 (moisture–dynamics response).
- Synergy: Combined-insertion percentage improvement minus the larger temperature-only or wind-only improvement over lead days 40–60. Positive values indicate additional benefit; negative values indicate worse performance than the stronger single-variable experiment.
- Z500 synergy (Fig. A5): FourCastNet V2: +11.6; Aurora: −7.0; Prithvi WxC: +4.6; GraphCast: −28.4 percentage points.
- Q850 synergy (Fig. A5): FourCastNet V2: +2.5; Aurora: +11.4; Prithvi WxC: +3.2 percentage points..

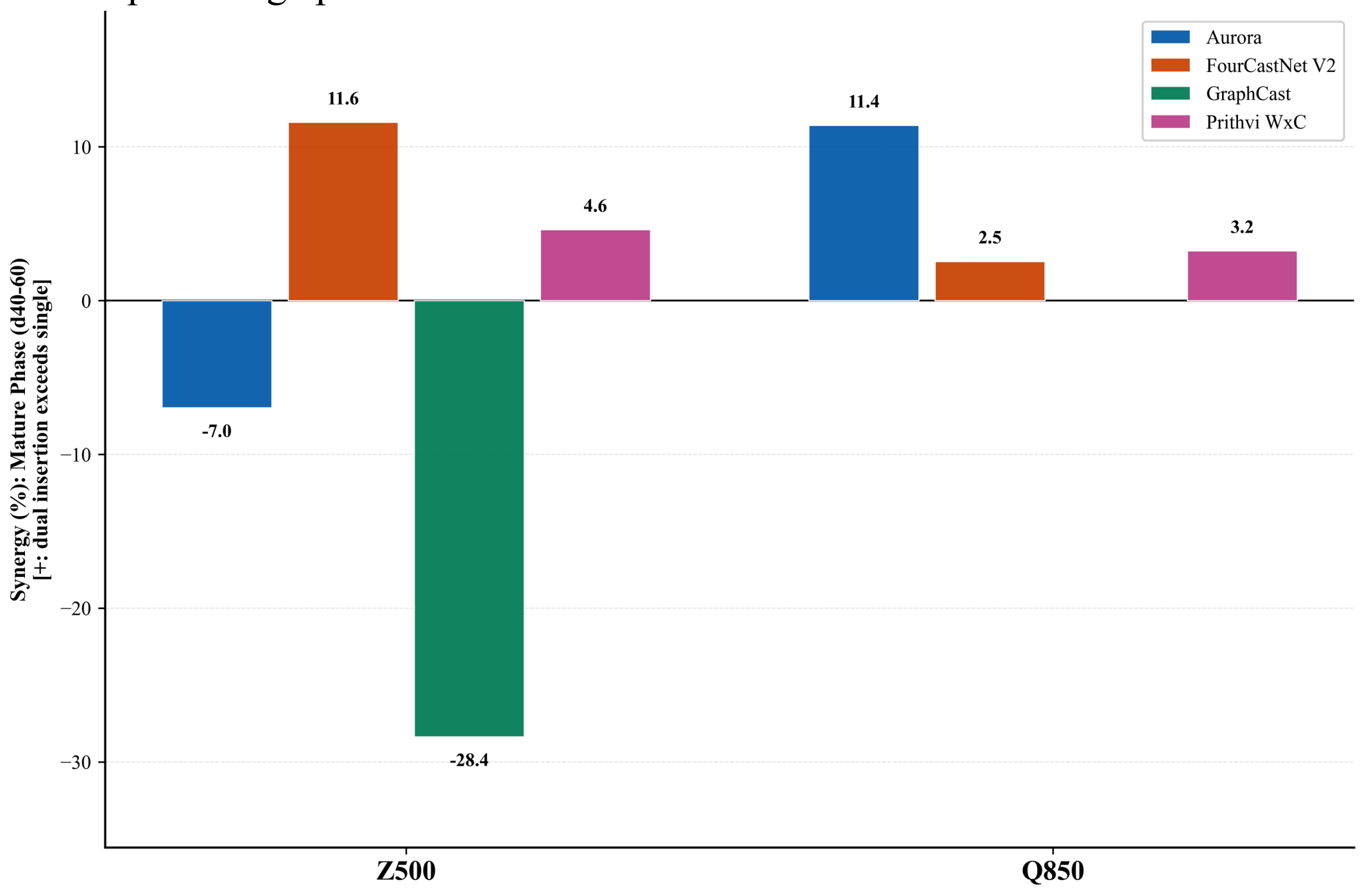

**Fig. A5.** TUV synergy bar chart, including GraphCast's Z500 synergy as a fourth bar (Q850 unavailable for GraphCast).

Fig. A5 shows these four synergy values directly, with FourCastNet V2 and Prithvi WxC positive on both pathways, Aurora mixed, and GraphCast's Z500 synergy strongly negative.

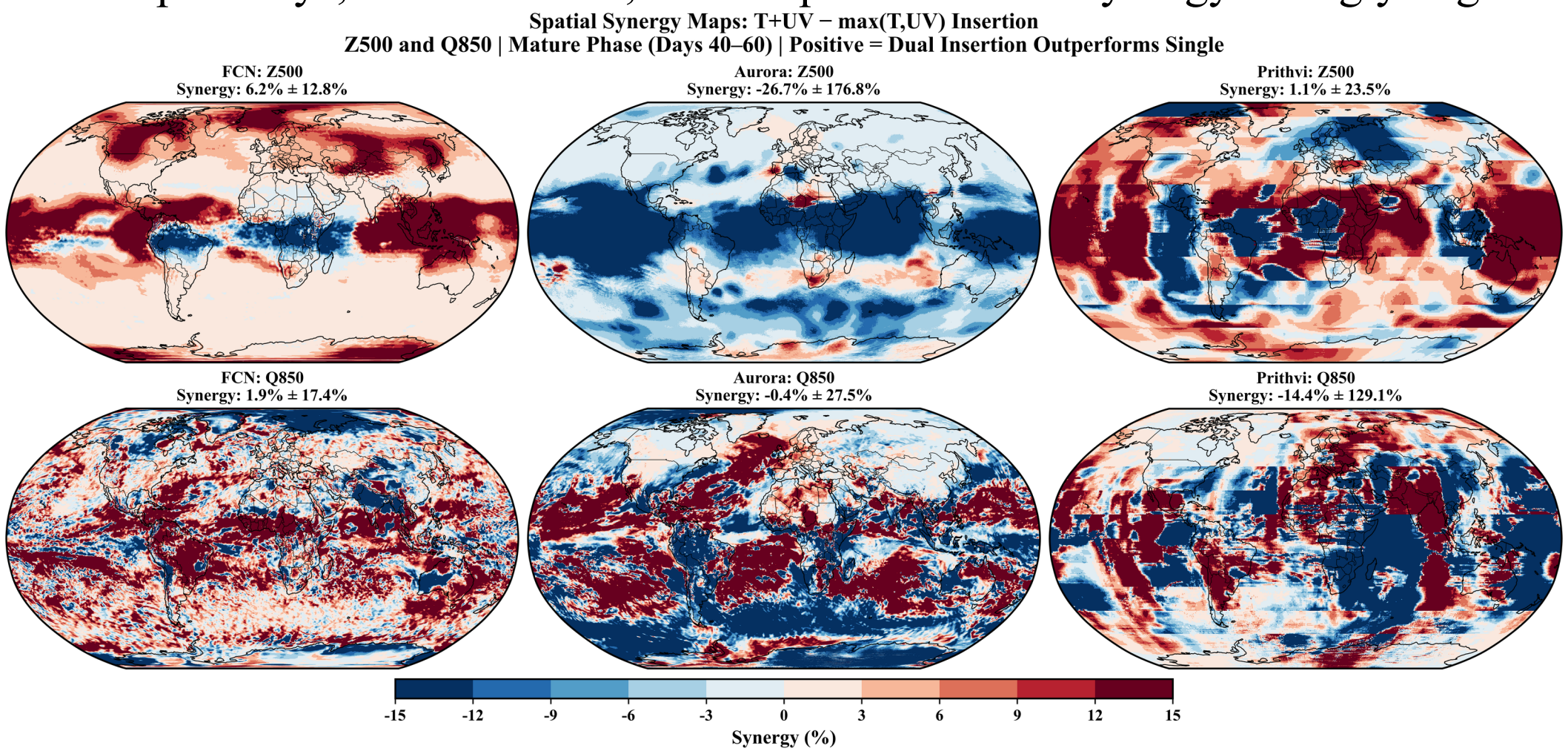

**Fig. A6.** TUV synergy spatial maps. Neither Prithvi WxC's nor GraphCast's combined-insertion data include the per-gridpoint spatial fields needed to compute this figure.

Fig. A6 shows the spatial structure behind these synergy scores for FourCastNet V2 and Aurora, the only two models for which per-gridpoint TUV-synergy data exist.

## Appendix B: Supplementary Cross-Variable Diagnostics

### B1. Cloud-Fraction Regime Analysis

Cloud regimes are diagnostic masks, not insertion regions: the inserted variable is corrected globally from day 30, and forecast responses are stratified by reference-state cloud classes: clear, cirrus, mixed, and opaque. This evaluates the same global intervention across local dynamical and moist-physical regimes. Cloud cover also indicates atmospheric conditions: clear skies often accompany subsidence, radiative cooling, and quiescent flow; opaque clouds commonly accompany organized ascent, fronts, or deep convection; and mixed clouds often mark transitional or baroclinically active environments. This stratification tests whether models propagate inserted information consistently across balanced large-scale flow and moist, cloud-active regimes.

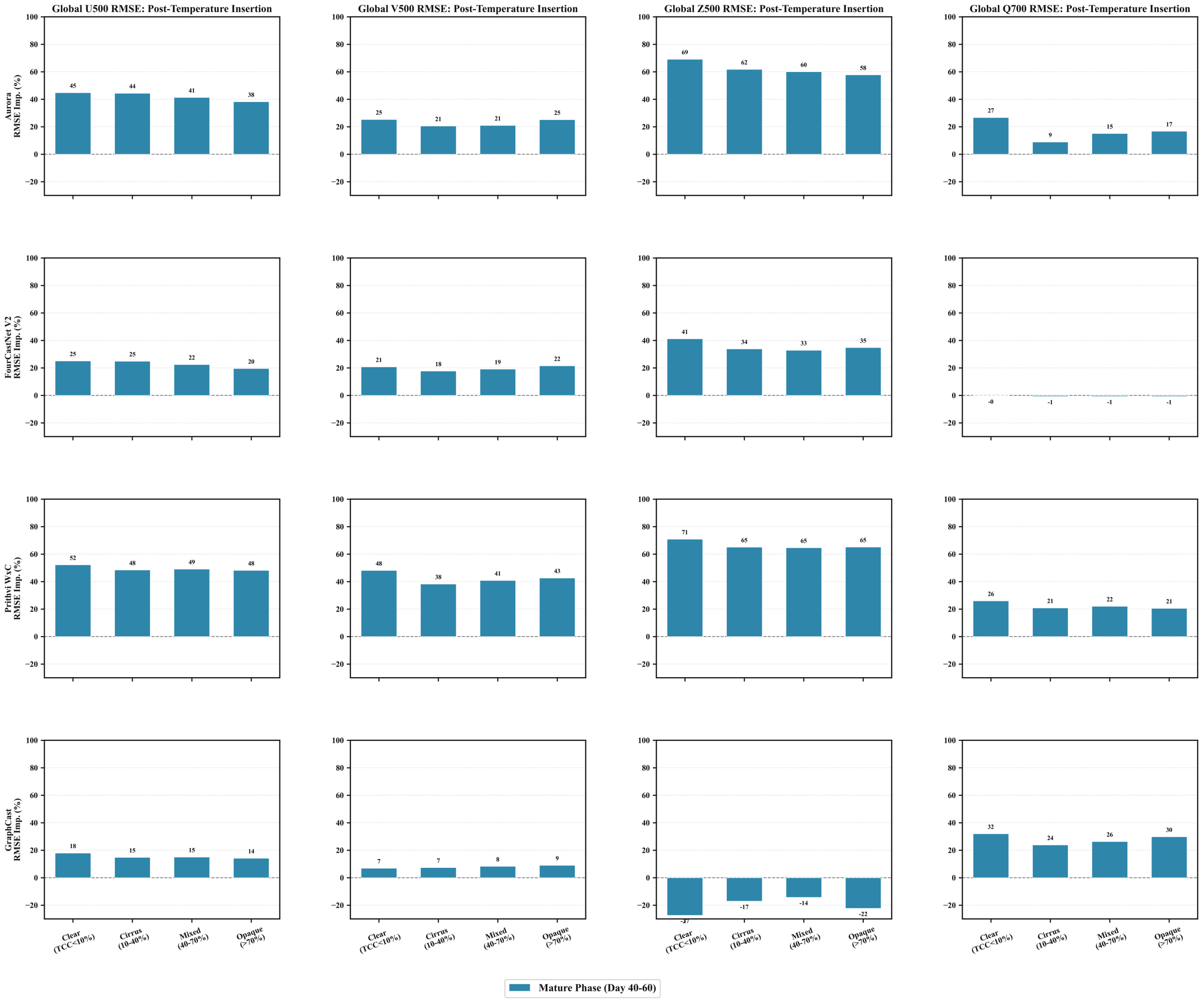

**Fig. B1.** Global Temperature Insertion results stratified by cloud regime across 4 models. GraphCast's Q700 panel is approximated from its 850-hPa humidity data, since GraphCast's humidity fields do not include a 700-hPa level.

- **FCN - skill tracks dynamics rather than cloud amount:** Improvements peak in Mixed conditions for four diagnostics: Z500 and U500 under temperature insertion, and T500 and Z500 under wind insertion. Weaker Clear and Opaque responses produce an inverted-U pattern, consistent with stronger thermal-wind coupling in frontal/baroclinic regions with steep temperature gradients and vertical shear. Cloud amount alone therefore does not explain correction skill.

- **Aurora and GraphCast - moisture coupling weakens in transitional regimes:** Aurora's wind-insertion Q850 benefit peaks in Clear and Opaque conditions and dips in Cirrus and Mixed; temperature-insertion Q700 shows a weaker version. This suggests coupling may be easier within a dominant physical regime than amid mixed processes. GraphCast shows a similar dip and recovery under temperature insertion: Clear 32.0%, Cirrus 23.9%, Mixed 26.3%, and Opaque 29.8%. Its Q850 serves as a Q700 proxy (Fig. B1), limiting direct comparison. This two-model pattern is absent in FCN and Prithvi.

- **Prithvi WxC - consistently positive moisture coupling:** Temperature insertion improves Q700 across Clear (26.0%), Cirrus (20.8%), Mixed (22.0%), and Opaque (20.5%) conditions. The narrow, positive range indicates relatively weak regime sensitivity, with no sign reversal; explaining the smaller differences requires further analysis.

## B2. Terrain Regime Analysis

Insertion is applied globally, with forecast responses evaluated using masks for low-, moderate-, and high-elevation land, ocean, and polar regions. This compares the same correction across surface and orographic conditions.

Terrain modifies atmospheric balance. Hydrostatic and geostrophic relationships are often more directly expressed over flat land and ocean, while complex terrain introduces mountain waves, slope circulations, boundary-layer effects, orographic lifting, lee-side pressure perturbations, and ageostrophic motion. Terrain stratification tests whether learned balance relationships hold under topographic forcing.

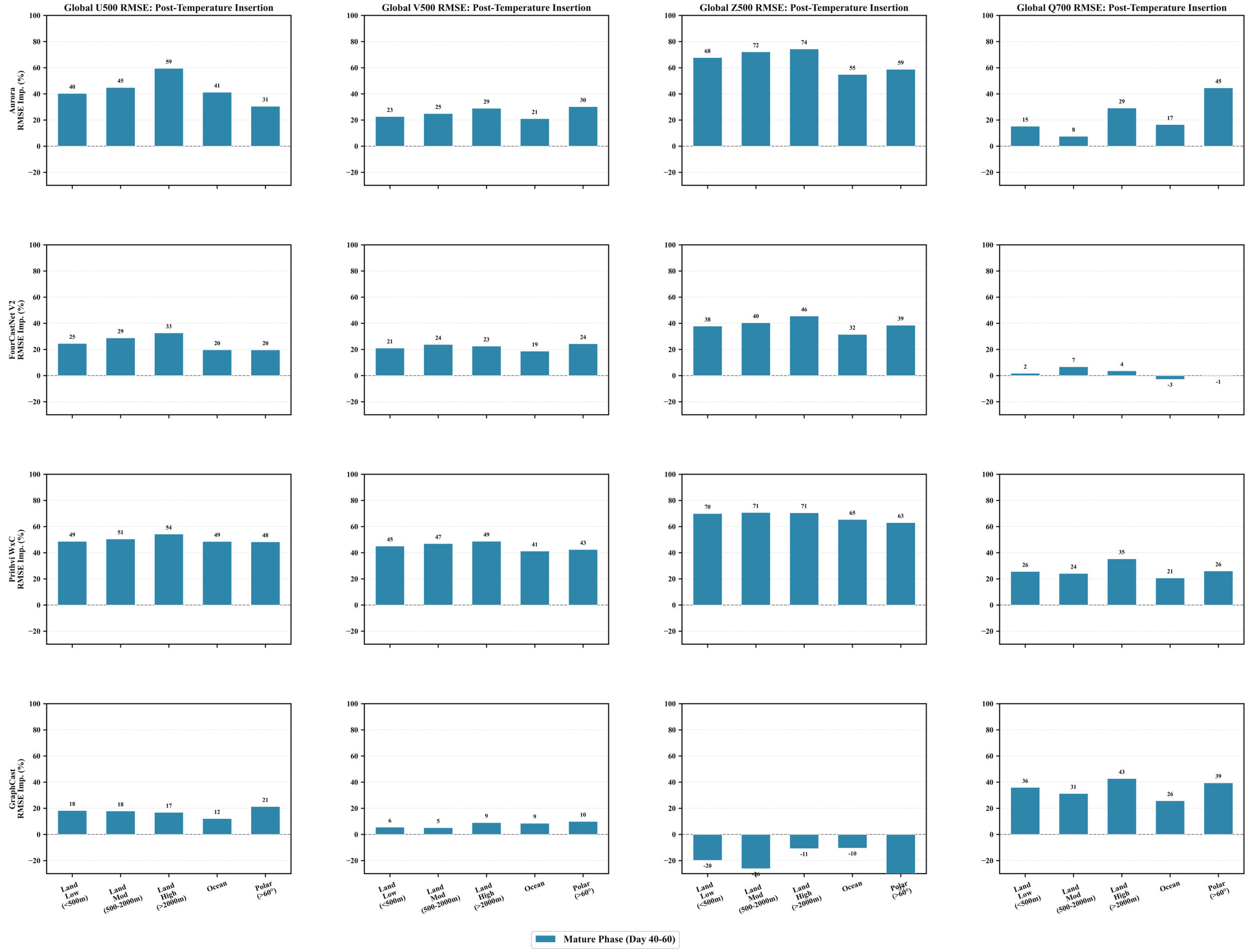

**Fig. B2.** Temperature Insertion results stratified by terrain.

Fig. B2 shows this stratification directly for temperature insertion, with each panel giving the mature-phase RMSE improvement across the five terrain classes for one model.

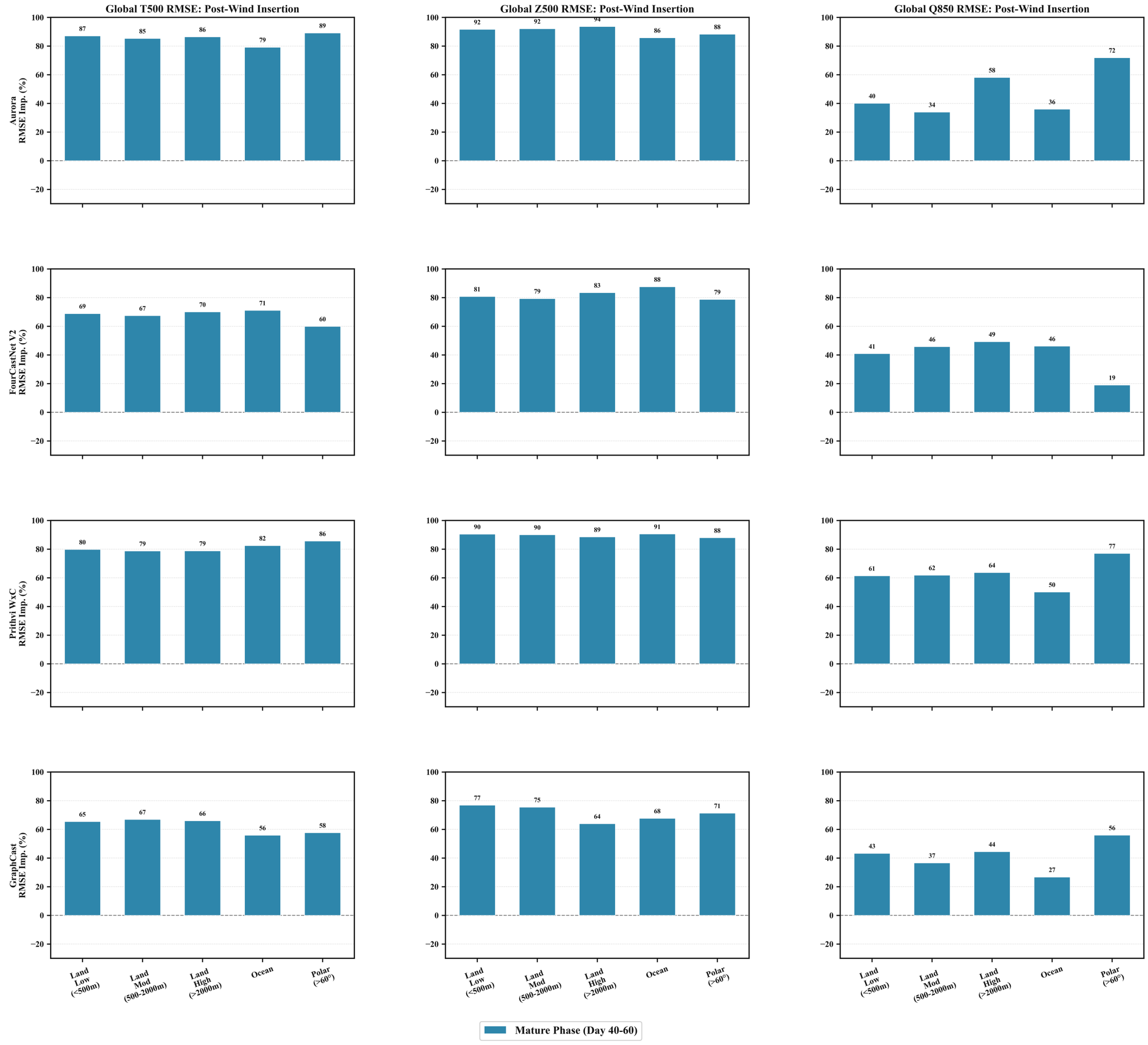

**Fig. B3.** Wind Insertion results stratified by terrain.

- **Prithvi WxC: no distinct high-terrain breakdown.** Under temperature insertion, Land High has the strongest V500 improvement (+48.7%). Under wind insertion, its Z500 improvement (+88.5%) is comparable to other classes (88.0% to 90.6%). Orographic forcing therefore does not uniquely disrupt its learned balance relationships.

- **Aurora: polar peaks remain inconclusive.** T500 and Q850 improvements peak at the Pole, consistent with stronger Coriolis control. However, Polar T500 has the largest baseline error, while Polar Q850 has the smallest. These different baseline conditions complicate interpretation; one deterministic rollout without an ensemble and five class comparisons also leaves chance as an explanation.

- **FCN: polar moisture weakness contrasts with Aurora.** Despite the largest baseline error and thus substantial room for improvement, FCN shows its weakest moisture benefit at the Pole. Baseline magnitude alone does not explain this pattern. It may indicate greater reliance on smaller-scale moisture structure, whereas Aurora may exploit large-scale polar advection.

- **Prithvi WxC: strong wind-to-temperature coupling across terrain.** T500 improvements span 78.7% to 85.7%, with Ocean second-highest (+82.5%) behind Polar (+85.7%). Strong performance therefore extends beyond flat land.

- **Aurora: polar Z500 weakness is less attributable to baseline size.** Ocean has the smallest Z500 baseline error, potentially limiting improvement, but Polar is similarly weak despite the largest baseline. FCN shows no corresponding polar weakness. Unequal class sizes and five-way comparisons leave sampling variability unresolved; another realization or significance test is needed to establish a robust pattern.

### B3. Coupling Score

As a single cumulative summary of everything examined in this study, each model's own coupling matrix: one number per (inserted variable → response variable) pair, condensing the full insertion experiment into a single cross-variable score.

What it measures: for every insertion experiment (temperature, wind, or geopotential forced to ERA5/MERRA-2 truth), how much forcing that one field corrects every other, non-inserted field averaged over all pressure levels and over the mature phase (day 30-60) of the rollout.

Geopotential height is included here as a third forced variable alongside temperature and wind purely to complete this coupling-score diagnostic (Fig. B4); unlike temperature and wind, it is not a quantity directly measured by any real observing system, so inserting it is not presented elsewhere in this study as a standalone OSSE experiment in its own right.

$$C(A \rightarrow B) = 100 \times [\ \mathrm{RMSE}_{ctrl}(B) - \mathrm{RMSE}_{ins,A}(B)\ ] \ /\ \mathrm{RMSE}_{ctrl}(B)$$

where A = the variable forced to truth, B = the response variable being measured, $\mathrm{RMSE}_{ctrl}(\mathrm{B})$ = free-running control RMSE of B vs. truth, $\mathrm{RMSE}_{ins,A}(\mathrm{B})$ = RMSE of B vs. truth under insertion experiment A, both averaged over all pressure levels and the mature-phase window before the ratio is taken. Positive C(A→B) means inserting A improves B; the diagonal is ~100% and excluded as trivial (self-forced).

GraphCast's Specific Humidity column is populated (28.8% for temperature insertion, 32.9% for wind insertion, 15.2% for geopotential insertion); the matrix is fully populated for all four models with no missing cells. The full matrix is summarized in the text above (Sections 4.2-4.4, 6) and is not reproduced as a table here since the underlying figure (Fig. B4) already presents it in full.

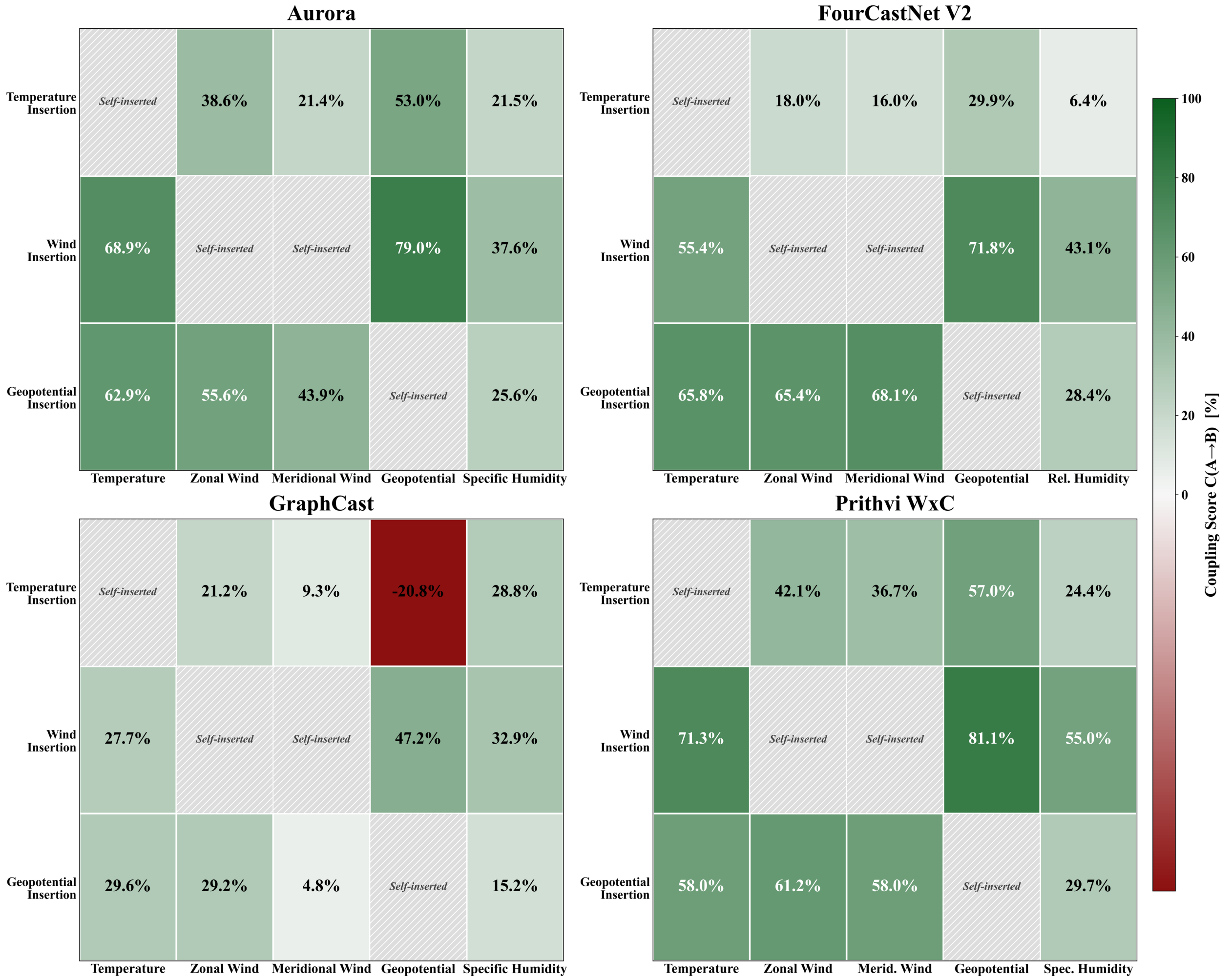

**Fig. B4.** Cross-variable coupling matrix, all 4 models.

Fig. B4 shows this matrix in full for all four models, condensing every insertion-to-response pathway examined in this study into one figure.